\documentclass[letterpaper, 10 pt, conference]{ieeeconf}  

\IEEEoverridecommandlockouts                              

\usepackage{amsmath,amsthm,amssymb,bm}
\usepackage{graphicx,fixltx2e}
\usepackage{hyperref}
\usepackage{xy}
\usepackage[linesnumbered,ruled,vlined]{algorithm2e}
\usepackage{bbm}
\usepackage{todonotes}
\usepackage{float}
\usepackage{cite}
\usepackage{cleveref}

\newcommand{\N}{\mathbb{N}}

\newcommand{\R}{\mathbb{R}}

\newcommand{\tbx}{\textbf{x}}

\let\oldnl\nl
\newcommand{\nonl}{\renewcommand{\nl}{\let\nl\oldnl}}

\makeatletter
\newcommand{\pushright}[1]{\ifmeasuring@#1\else\omit\hfill$\displaystyle#1$\fi\ignorespaces}
\newcommand{\pushleft}[1]{\ifmeasuring@#1\else\omit$\displaystyle#1$\hfill\fi\ignorespaces}
\usepackage{hyperref}
\makeatletter
\def\@citex[#1]#2{\leavevmode
\let\@citea\@empty
\@cite{\@for\@citeb:=#2\do
{\@citea\def\@citea{,\penalty\@m\ }%
\edef\@citeb{\expandafter\@firstofone\@citeb\@empty}%
\if@filesw\immediate\write\@auxout{\string\citation{\@citeb}}\fi
\@ifundefined{b@\@citeb}{\hbox{\reset@font\bfseries ?}%
\G@refundefinedtrue
\@latex@warning
{Citation `\@citeb' on page \thepage \space undefined}}%
{\@cite@ofmt{\csname b@\@citeb\endcsname}}}}{#1}}
\makeatother

\newenvironment{theorem}[2][Theorem]{\begin{trivlist}
\item[\hskip \labelsep {\bfseries #1}\hskip \labelsep {\bfseries #2.}]}{\end{trivlist}}

\allowdisplaybreaks

\title{\LARGE \bf
Real-Time Tube-Based Non-Gaussian Risk Bounded Motion Planning for Stochastic Nonlinear Systems in Uncertain Environments via Motion Primitives
}

\author{Weiqiao Han$^{*1}$, Ashkan Jasour$^{*2}$, Brian Williams$^{1}$
\thanks{$^{1}$Computer Science and Artificial Intelligence Laboratory, Massachusetts Institute of Technology, \{weiqiaoh,williams\}@mit.edu.
$^{2}$ Massachusetts Institute of Technology, jasour@mit.edu. Now with NASA Jet Propulsion Lab, California Institute of Technology.
$^{*}$These authors contributed equally to the paper.}%
}

\begin{document}

\newcounter{eqn}

\maketitle
\thispagestyle{empty}
\pagestyle{empty}

\begin{abstract}
We consider the motion planning problem for stochastic nonlinear systems in uncertain environments. 
More precisely, in this problem the robot has stochastic nonlinear dynamics and uncertain initial locations, and the environment contains multiple dynamic uncertain obstacles. Obstacles can be of arbitrary shape, can deform, and can move. All uncertainties do not necessarily have Gaussian distribution.
This general setting has been considered and solved in \cite{han2022non}. 
In addition to the assumptions above, in this paper, we consider long-term tasks, where the planning method in \cite{han2022non} would fail, as the uncertainty of the system states grows too large over a long time horizon.
Unlike \cite{han2022non}, we present a real-time online motion planning algorithm.  
We build discrete-time motion primitives and their corresponding continuous-time tubes offline, so that almost all system states of each motion primitive are guaranteed to stay inside the corresponding tube. 
We convert probabilistic safety constraints into a set of deterministic constraints called risk contours. 
During online execution, we verify the safety of the tubes against deterministic risk contours using sum-of-squares (SOS) programming. 
The provided SOS-based method verifies the safety of the tube in the presence of uncertain obstacles without the need for uncertainty samples
and time discretization in real-time.
By bounding the probability the system states staying inside the tube and bounding the probability of the tube colliding with obstacles, our approach guarantees bounded probability of system states colliding with obstacles. 
We demonstrate our approach on several long-term robotics tasks.
\end{abstract}

\section{Introduction}
When a robot moves around in the real world, it encounters all kinds of stochasticity from nature. The uneven terrain, the volatile weather, and surrounding uncertain moving obstacles all could pose challenges to the robot navigation. 
How to plan motions for the robot to navigate safely and reach the goal is a long-standing problem. 

Traditional planning algorithms, such as rapidly exploring random tree (RRT), probabilistic roadmap (PRM), and virtual potential field methods, plan paths in deterministic environments.
Planning algorithms assuming stochasticity and uncertainty are mainly limited to Gaussian uncertainty and convex obstacles \cite{axelrod2018provably, schwarting2017parallel, dai2019chance, lew2020chance, dawson2020provably, luders2010chance, blackmore2009convex}.
There are also sampling based methods that do not necessarily assume Gaussian uncertainty \cite{blackmore2010probabilistic, janson2018monte, calafiore2006scenario, cannon2017chance}.

\cite{han2022non} considered the most general setting so far, where (a) the system dynamics is stochastic and nonlinear, and the uncertainty is not necessarily Gaussian; (b) the initial position of the system is uncertain and is necessarily Gaussian; (c) the obstacles can be of arbitrary shape, can deform, can move, and have arbitrary uncertainty, not necessarily Gaussian. \cite{han2022non} proposed a trajectory optimization method to plan a path with bounded risk, where risk is defined to be the probability of colliding with obstacles or not reaching the goal. However, since the planned path is open loop, the uncertainty grows as the system evolves. Therefore, the planning horizon is limited. We want to plan longer horizons for the robot to accomplish more complex tasks.
\begin{figure}[t!]
    \centering
    \includegraphics[width=0.48\textwidth]{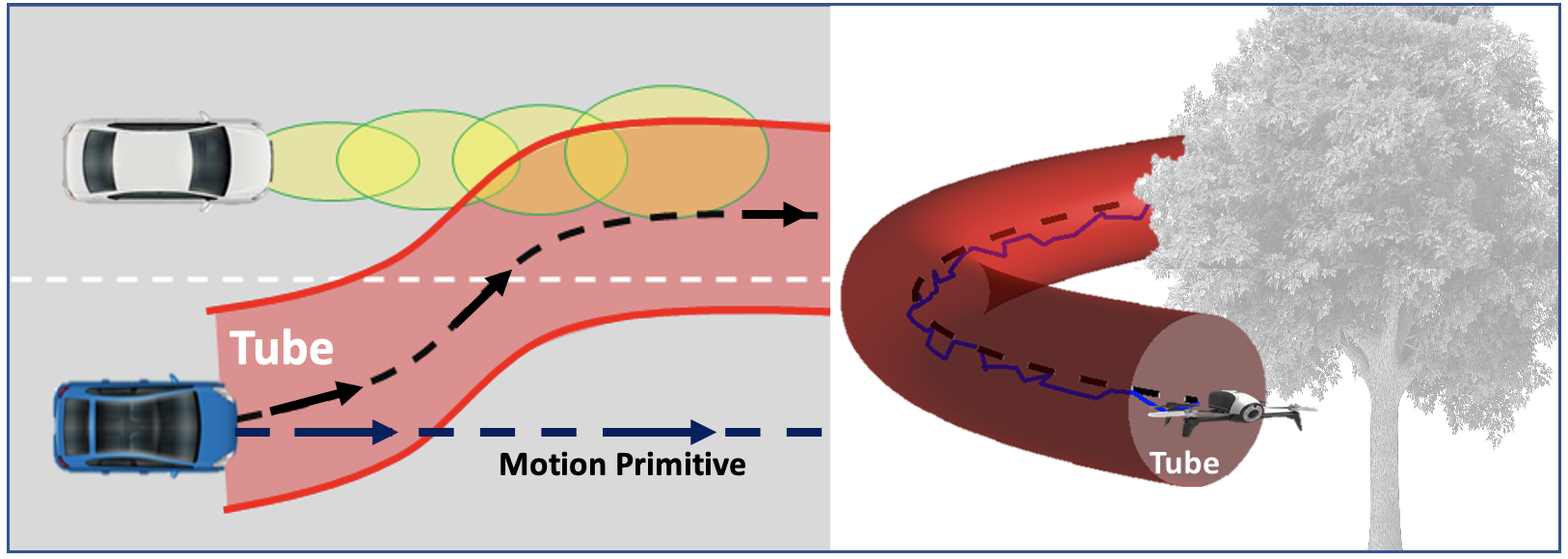}
    \caption{Motion primitives and tubes}
    \label{fig:tube_1}
\end{figure}

One would think of using closed loop control to track the nominal trajectory planned by the trajectory optimization. However, currently closed loop controllers are usually designed for linear systems. For example, in \cite{okamoto2019optimal}, controllers are designed to steer the stochastic time-varying linear system to the goal with desired covariance, while satisfying chance constraints. It is not obvious how such controller would behave on a linearized stochastic nonlinear system. It is also unlikely to incorporate the design of such closed loop controller into the trajectory optimization for long-term motion planning.

One approach to long-term tasks is online planning with motion primitives \cite{howard2008state,howard2007optimal,yang2019online,bottasso2008path,low2021prompt,paraschos2013probabilistic}. For example, \cite{howard2008state} considers state space sampling of motion primitives, and then use numerical methods to solve for the controller. They assume the system dynamics is deterministic. In contrast, we are going to use motion primitives for stochastic systems. 

In this paper, we consider the motion planning problem in the same setting as in \cite{han2022non}, satisfying conditions (a)--(c) listed above, and in addition, we consider (d) long-term tasks, where the short-term planning method in \cite{han2022non} would fail, as the uncertainty of the system states grows too large over a long time horizon.
We present a real-time online motion planning approach to this general problem. 
We first generate motion primitives in control space over a discrete time horizon, and then build a continuous-time tube in state space for each motion primitive, such that almost all system states remain in the tube over the discrete time horizon. 
We provide theoretical guarantees that the probability of system states outside of the tube is bounded above by a very small number.
For the uncertain environment, we use concentration inequalities to convert the uncertain obstacles into deterministic risk contours. 
During online execution, we verify the safety of the tubes against deterministic risk contours using sum-of-squares (SOS) programming. 
Among all feasible tubes that stay inside the low-risk contour, an objective is used to score their corresponding motion primitives, and the one with the highest score is selected and executed.
Our approach bridges the gap between the low-level discrete-time control sequence planning of stochastic nonlinear systems, and the high-level continuous-time state space planning, where safety can be verified via SOS programming. 
Our approach provides theoretical guarantees that the probability of system states colliding with any obstacle is bounded above by a small number. 
We demonstrate our approach on several long-term robotics tasks.

\section{Problem Definition}
Consider the stochastic nonlinear dynamical system represented by 
\begin{equation}
    \mathbf{x}_{t+1} = f(\mathbf{x}_t, \mathbf{u}_t, \mathbf{\omega}_t)\label{eq:sysdyn}
\end{equation}
where $\mathbf{x}_t \in \mathcal{X}, \mathbf{u}_t \in \mathcal{U}$ are the state and control inputs at time $t$, respectively, $\mathcal{X} \subseteq \R^n$ is the state space, $\mathcal{U} \subseteq \R^m$ is the control space, $n,m \in \mathbb{N}$, $\mathbf{\omega}_t$ is the noise at time $t$ and it may not necessarily have Gaussian distribution.
The initial state $\mathbf{x}_0$ can either be deterministic, or can be a random variable with some known probability distribution, not necessarily Gaussian. 
In the environment, there are multiple obstacles, denoted by $\mathcal{O}_i, i = 1,\ldots, M$, where $M\in \N^+$, and each obstacle $\mathcal{O}_i$ has a polynomial representation
\begin{align}\label{eq:obs}
    \mathcal{O}_i(\tilde{w}_i, t) = \{\mathbf{x} \in \mathcal{X} : \  p_i(\mathbf{x}, \tilde{\omega}_i, t) \leq 0  \}, \ i=1,...,M
\end{align}
where $p_i$ is a polynomial and $\tilde{\omega}_i$ is a random variable with some known probability distribution, not necessarily Gaussian. 
The obstacle can change its position and shape over time, and hence the time variable $t$ in the polynomial representation.
Note that we assume the future trajectory of any obstacle is known with known noise. 
This assumption is valid, because there can be a prediction module in the system that predicts future trajectories of surrounding obstacles. 
In fact, in the autonomous driving community, there is a whole research area devoted to predicting future trajectories of agents, including vehicles, bicyclists, and pedestrians, in a traffic scene \cite{gu2021densetnt, lu2022kemp}. 
We define the risk to be the probability of collision with any uncertain obstacle at any time step.
The goal is to plan control sequences $\mathbf{u}_0, \mathbf{u}_1, \ldots$ online to steer the system into the goal region $\mathbf{x}_{goal}\subseteq \mathcal{X}$. More precisely, we want to solve the following planning problem:

\noindent\textbf{Risk Bounded Motion Planning Problem}
\begin{equation}
\begin{aligned}
\min_{\mathbf{u}} \quad &E[l_f(\mathbf{x}_T) + \sum_{t=0}^{T-1} l(\mathbf{x}_t, \mathbf{u}_t,\omega_t)]\\
\text{s.t.} \quad & \mathbf{x}_{t+1} = f(\mathbf{x}_t, \mathbf{u}_t, \omega_t), \  \ |_{t = 0}^{T-1},\\
& \text{Prob}(\mathbf{x}_t \in \mathcal{O}_i(\tilde{\omega}_i,t)) \leq \Delta_o, \ \  |_{t = 0}^{T-1}, \ |_{i=1}^{M}\\
& \text{Prob}(\mathbf{x}_T \not\in \mathbf{x}_{goal}) \leq \Delta_{goal},\\
& \mathbf{x}_0 \sim pr(\mathbf{x}_0)
\end{aligned}
\end{equation}
where $\Delta_o,\Delta_{goal} \in [0,1]$ are the given acceptable risk levels, and $pr(\mathbf{x}_0)$ is the given probability distribution of the initial system states. We assume that either there is a global planner, obtained from the method in \cite{jasour2021convex}, for example, which roughly traces out a general path from the initial position to the goal region, or there is a high-level objective function that guides the system to the goal region. 

\subsection{Notations and Definitions}
Let $\R[x]$ be the polynomial ring in the variables $\mathbf{x} = (x_1,\ldots,x_n)$ with real coefficients. 
A polynomial $p(\mathbf{x}) \in \R[x]$ can be written as $p(\mathbf{x}) = \sum_{\bm{\alpha}\in\N^n} p_{\bm{\alpha}} \mathbf{x}^{\bm{\alpha}}$, where $\bm{\alpha} = (\alpha_1,\ldots,\alpha_n) \in \N^n$ and $\mathbf{x}^{\bm{\alpha}} = \prod_{i=1}^n x_i^{\alpha_i}$ is a monomial in standard basis. 

Let $(\Omega, \Sigma, \mu)$ be a probability space, where $\Omega$ is the sample space, $\Sigma$ is the $\sigma$-algebra of $\Omega$, and $\mu: \Sigma \to [0,1]$ is the probability measure on $\Sigma$. 
Suppose $\mathbf{x} \in \Omega \subseteq \R^n$ is an $n$-dimensional random vector. 
Let $\bm{\alpha} = (\alpha_1,\ldots,\alpha_n)\in \N^n$.
The expectation of $\mathbf{x}^{\bm{\alpha}}$ defined as $m_{\bm{\alpha}} = E[\mathbf{x}^{\bm{\alpha}}] $ is a {moment of order $\alpha$}, where $\alpha = \sum_i \alpha_i$.
The sequence of all moments of order $\alpha$, denoted by $\mathbf{m}_\alpha$, is the expectation of all monomials of order $\alpha$ sorted in graded reverse lexicographic order (grevlex).

\textbf{Sum-of-Squares Polynomial:} 
Polynomial {\small$\mathcal{P}(\mathbf{x})$} is a sum-of-squares polynomial if it can be written as a sum of {finitely} many squared polynomials, i.e., {$\mathcal{P}(x)= \sum_{j=1}^{m} p_j(\mathbf{x})^2$} for polynomials $p_j(\mathbf{x})$, $1\leq j\leq m$. 
Checking if a polynomial is SOS can be cast into a semidefinite programming program.
One can use the packages like Yalmip \cite{Yalmip_1} and Spotless \cite{Spot_1} to check if a polynomial is SOS.

\section{Method}
In this section, we first review how to convert uncertain obstacles in the environment into risk contours \cite{jasour2021real}. Second, we review how to verify the safety of the tube using risk contours and SOS programming \cite{jasour2021real}. Next we present how to generate motion primitives and their corresponding tubes, and discuss several points to optimize implementation in online execution. Finally, we prove the theoretical guarantees on the bounded risk. 
\subsection{Risk Contour Representation of Uncertain Environments}\label{sec:method_1}
In this subsection, we review risk contours \cite{Contour1}, which have been used in \cite{han2022non, jasour2021convex, jasour2021real}.

Suppose we want to bound the risk of colliding with the obstacle $\mathcal{O}$ defined in \eqref{eq:obs} by $\Delta$, i.e., $\text{Prob}(\tbx \in \mathcal{O}) = \text{Prob}(p(\mathbf{x}, \tilde{\omega}, t) \leq 0) \leq \Delta$. Then, the associated risk contour is defined as 
\begin{equation} \label{cc_3}
  \mathcal{C}_{r}^{\Delta}(t)= \{\mathbf{x} \in \mathcal{X}: \hbox{Prob}\left( p(\mathbf{x}, \tilde{\omega}, t) \leq 0   \right) \leq \Delta \}
\end{equation}

The following result holds true.

\begin{theorem}{(Theorem 1 in \cite{jasour2021real})}
The set

\begin{equation} \label{risk_contour}
  \hat{\mathcal{C}}_{r}^{\Delta}(t)= 
    \left\{\mathbf{x} \in \mathcal{X}: \frac{E[g^2] - E[g]^2}{E[g^2]} \leq\Delta, E[g] \leq 0,\right\}
\end{equation}
where $g=-p(\mathbf{x}, \tilde{\omega},t)$, 
is the inner approximations of the risk contour in \eqref{cc_3}.
\end{theorem}
In the proof of the Theorem above we use concentration
inequalities to provide bounds on random variables \cite{jasour2021real}. 
More precisely, by Cantelli’s inequality, for any random variable $z$,
\begin{align} \label{eq:concentration_ineq}
    \text{Prob}(z \geq 0) \leq \frac{E[z^2] - E[z]^2}{E[z^2]},
\end{align} 
whenever $E[z] \leq 0 $. Any upper bound on the right-hand side of (\ref{eq:concentration_ineq}) would be an upper bound of the probability on the left-hand size.

The set $\hat{\mathcal{C}}_{r}^{\Delta}(t)$ is a region in state space where the probability of collision with the uncertain obstacle is $\leq\Delta$. 
For one obstacle, $\hat{\mathcal{C}}_{r}^{\Delta}(t)$ is defined by 2 polynomials. 
In general, for $M$ obstacles, the risk-bounded set $\hat{\mathcal{C}}_{r}^{\Delta}(t)$ is defined by $2M$ polynomials, 2 for each obstacle, involving $E[g_i]$ and $E[g_i^2]$, where $g_i = -p_i(\mathbf{x}, \tilde{\omega},t)$, $i=1,\ldots,M$, and $\hat{\mathcal{C}}_{r}^{\Delta}(t)$ is the region in state space where the probability of collision with any of the $M$ uncertain obstacles is $\leq\Delta$. 

\subsection{Tube Verification}\label{sec:method_2_tube_verification}
Here we briefly review tube safety verification. The provided method verifies the safety of the tube in the presence of uncertain obstacles without the need for uncertainty
samples and time discretization. Curious readers are suggested to refer to \cite{jasour2021real} for more details. Given a polynomial trajectory $\mathcal{P}(t)$ parameterized by $t$, a tube around it takes the form
\begin{equation}\label{tube}
\mathcal{Q}(\mathcal{P}(t)) =\{ \mathbf{x} \in \mathbb{R}^{n}:  (\mathbf{x}-\mathcal{P}(t))^T{Q}(\mathbf{x}-\mathcal{P}(t))\leq 1 \}   
\end{equation}
where $Q\in \mathbb{R}^{n \times n}$ is the given positive definite matrix and $t\in[t_0,t_f]$. 

We can use SOS programming to verify if a tube is inside the risk-bounded set $\hat{\mathcal{C}}_{r}^{\Delta}(t)$ in \eqref{risk_contour} over the entire planning time horizon $t\in[t_0,t_f]$. This is the following theorem. 

\begin{theorem}{(Theorem 3 in \cite{jasour2021real})}
Denote $P_{1i}(\tbx,t)={E}[g_i^2]$ and $P_{2i}(\tbx,t)={E}[g_i]$. 
The given tube $\mathcal{Q}(\mathcal{P}(t))$ over $t\in[t_0,t_f]$ is inside the risk-bounded set $\hat{\mathcal{C}}_{r}^{\Delta}(t)$ in \eqref{risk_contour} if the polynomials of $\hat{\mathcal{C}}_{r}^{\Delta}(t)$ take the following SOS representation:
\begin{align}\label{SOS_t_Cond1}
    & P^2_{2i}  \left(\mathcal{P}(t)+\hat{\mathbf{x}}_0,t\right)- (1-\Delta)P_{1i}(\mathcal{P}(t)+\hat{\mathbf{x}}_0,t)=\\
   & \quad\quad\quad\quad {\sigma_0}_i(t,\hat{\mathbf{x}}_0)+ {\sigma_1}_i(t,\hat{\mathbf{x}}_0)(t-t_0)(t_f-t) \notag\\ 
   & \quad\quad\quad\quad + {\sigma_2}_i(t,\hat{\mathbf{x}}_0)(1-\hat{\mathbf{x}}_0^T{Q}\hat{\mathbf{x}}_0) |_{i=1}^{M} \notag \\
    & P_{2i} (\mathcal{P}(t)+\hat{\mathbf{x}}_0,t)=\\
    & \quad\quad\quad\quad {\sigma_3}_i(t,\hat{\mathbf{x}}_0)+ {\sigma_4}_i(t,\hat{\mathbf{x}}_0)(t-t_0)(t_f-t)   \notag\\
    & \quad\quad\quad\quad + {\sigma_5}_i(t,\hat{\mathbf{x}}_0)(1-\hat{\mathbf{x}}_0^T{Q}\hat{\mathbf{x}}_0) |_{i=1}^{M}  \notag
\end{align}
where $\hat{\mathbf{x}}_0 \in \mathbb{R}^{n}$ is the variable vector, ${\sigma_j}_i(t,\hat{\mathbf{x}}_0),j=0,...,5, i=1,...,M$ are SOS polynomials with appropriate degrees. 
\end{theorem}
 As shown in the experiment section, one can verify the obtained safety SOS conditions in real-time. Note that the complexity of the provided safety SOS conditions is independent of the size of the planning time horizon $[t_0, t_f]$ and the length of the polynomial
trajectory  $\mathcal{P}(t)$. Hence, they can be easily used to verify the
safety of trajectories and their neighborhoods represented by tubes in uncertain environments over the long
planning time horizon.

\subsection{Motion Primitives and Their Corresponding Tubes}\label{sec:method_3_motion_primitives}
A motion primitive is defined to be a pre-computed sequence of control inputs $\{\mathbf{u}_0, \ldots, \mathbf{u}_{T-1}\}$ of length $T \in \N$.
When working with stochastic systems, each control $\mathbf{u}_i$ can depend on the state random variable $\mathbf{x}_i$, i.e., $\mathbf{u}_i=\mathbf{u}_i(\mathbf{x}_i)$.
In general, there are two ways to generate motion primitives for deterministic systems \cite{howard2008state}. One way is to sample the state space. In this way, the state constraints imposed by the environment, such as obstacle avoiding and goal reaching, can be easily satisfied. However, it is hard to satisfy dynamics constraints imposed by the underlying dynamical system. The other way is to sample the control space. The states are formed by simulating forward in dynamics. The state constraints in this case are not easily satisfied.
In light of this, we consider two ways to generate motion primitives and their associated tubes. One is based on control space sampling and the other is based on state space sampling.

\subsubsection{Method 1} Control space sampling. In this method, we first sample control sequences from the control space to get a desired motion primitive. Next, we fit a polynomial nominal trajectory $\bar{\tbx}_\theta(t)$, $t\in [0,1]$, parameterized by $\theta$ in some parameter space $\Theta$, and fit a polynomial tube to the motion primitive. Here, fitting a polynomial nominal trajectory rigorously means solving the following optimization problem
\begin{align*}
    \min_{\theta \in \Theta} & \sum_{k=1}^T |E[\mathbf{x}_k] - \bar{\tbx}_\theta(k/T)|^2\\
    \text{subject to} & \quad \text{system dynamics } (\ref{eq:sysdyn})\\
    & \quad \text{initial state distribution } \mathbf{x}_0 \sim p(\mathbf{x}_0)
\end{align*}
If $E[\mathbf{x}_k]$, $k = 1, \ldots, T$, can be calculated analytically or estimated using sampling, then the problem amounts to minimizing a polynomial function of $\theta$. 
Besides solving the optimization problem, we can also use sampling to approximately find a good enough $\theta$. 
The construction of the tube is to be explained in the next subsection. 

\subsubsection{Method 2} State space sampling. In this method, we first generate a desired polynomial nominal trajectory $\bar{\tbx}(t)$, $t\in [0,1]$, for some $\theta \in \Theta$. Next we apply trajectory optimization or sampling to look for the control sequence to follow the nominal trajectory so that the expected states are on or near the nominal trajectory. The trajectory optimization has the form of 
\begin{align*}
    \min_{\mathbf{u}_0,\ldots, \mathbf{u}_{T-1}} & \sum_{k=1}^T |E[\mathbf{x}_k] - \bar{\tbx}_\theta(k/T)|^2\\
    \text{subject to} & \quad \text{system dynamics } (\ref{eq:sysdyn})\\
    & \quad \text{initial state distribution } \mathbf{x}_0 \sim p(\mathbf{x}_0)
\end{align*}
If this optimization problem can be solved, we get a sequence of control inputs $\{\mathbf{u}_0,\ldots, \mathbf{u}_{T-1}\}$, which is the motion primitive.
Besides solving the optimization problem, we can also use sampling to approximately find a good enough control sequence $\{\mathbf{u}_0,\ldots, \mathbf{u}_{T-1}\}$.
Finally, we fit the trajectory with a tube, which is to be explained in the following subsection. 

\begin{figure}[t!]
    \centering
    \includegraphics[width=0.23\textwidth]{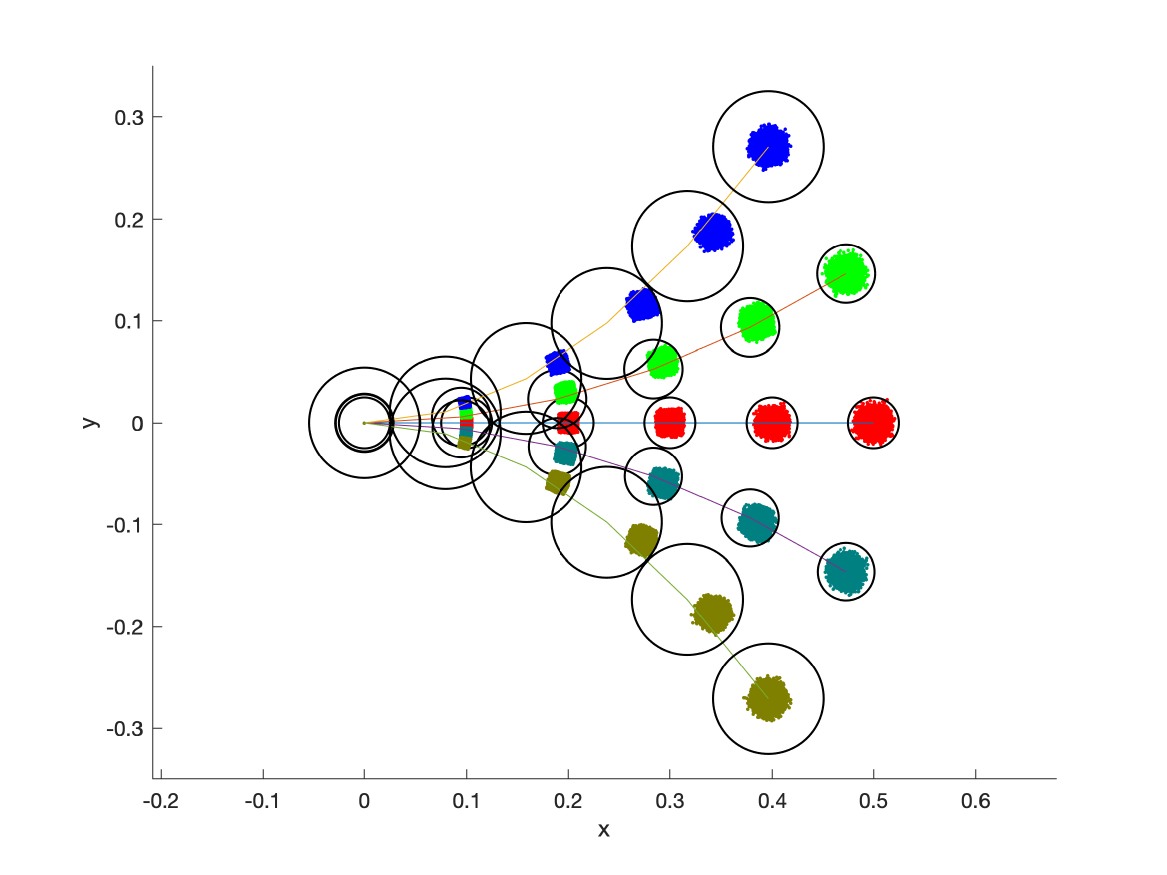}
    \includegraphics[width=0.23\textwidth]{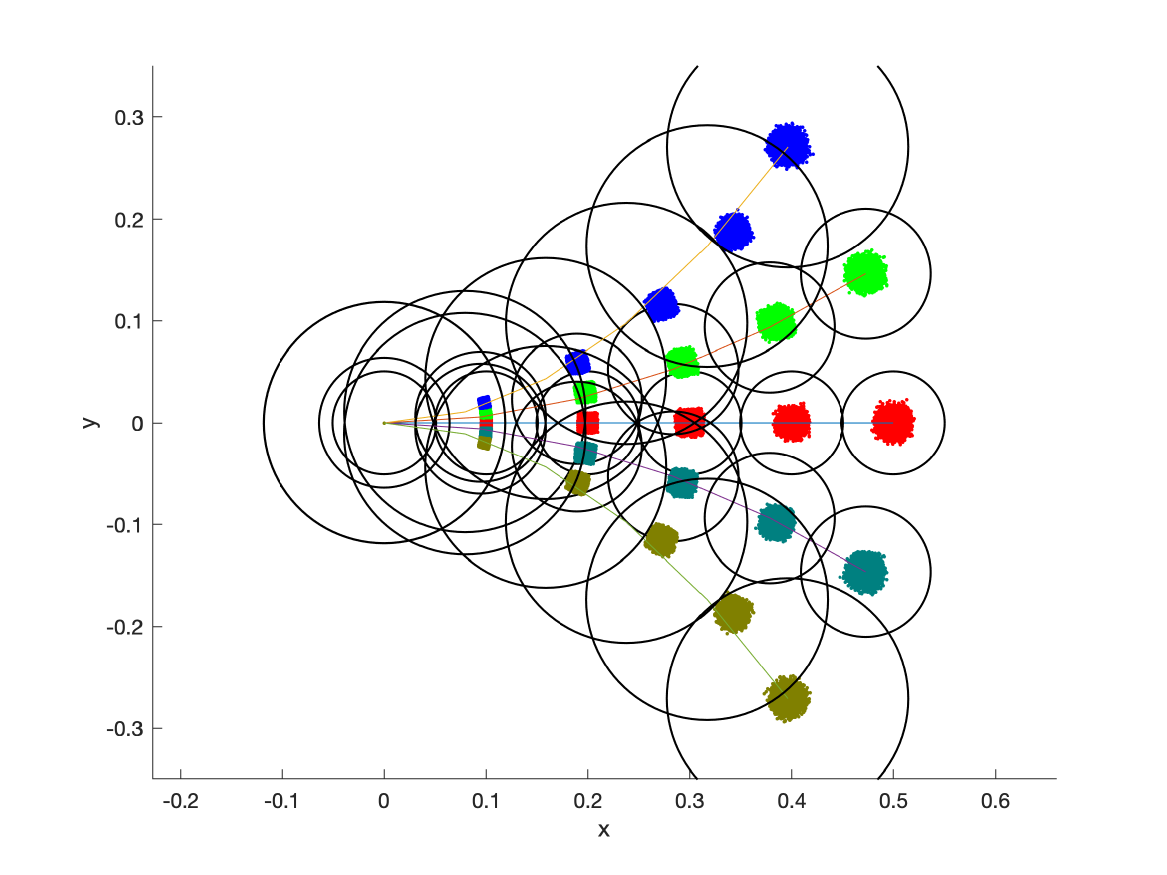}
    
    \includegraphics[width=0.23\textwidth]{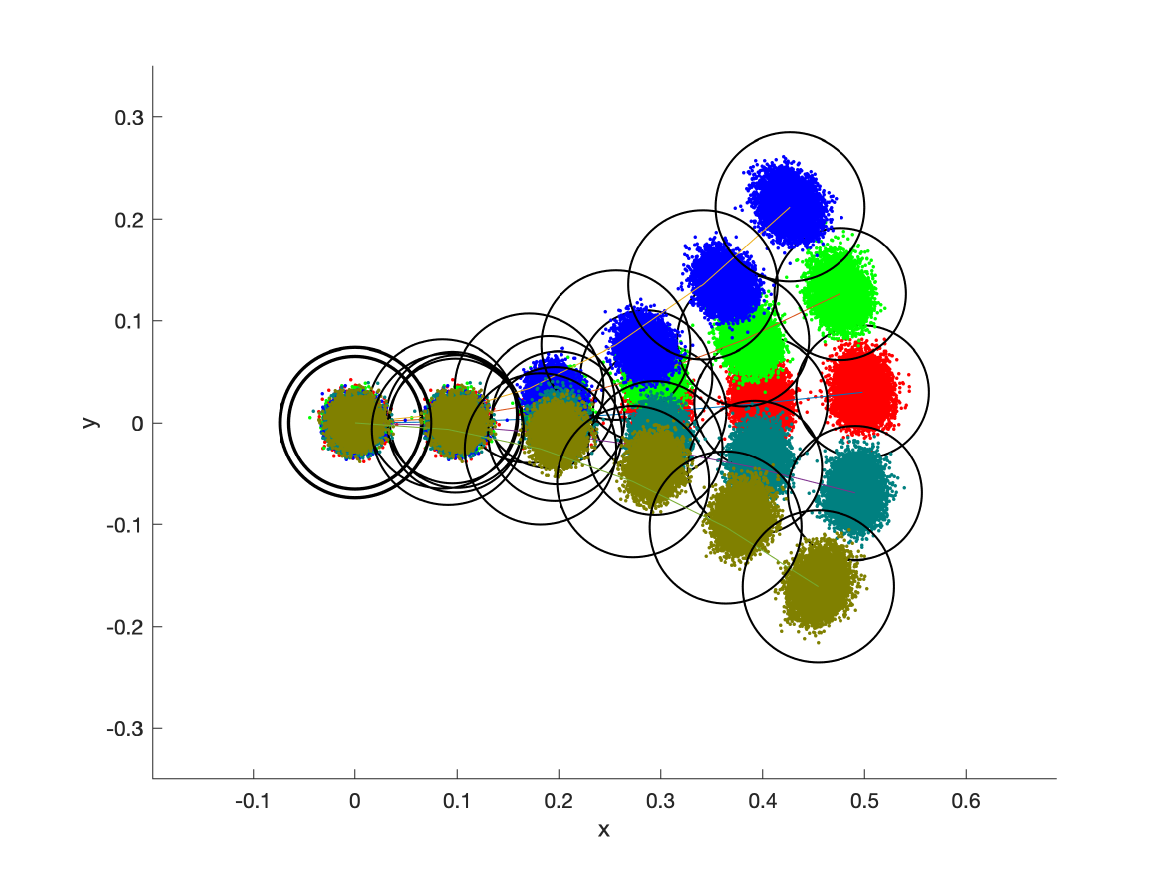}
    \includegraphics[width=0.23\textwidth]{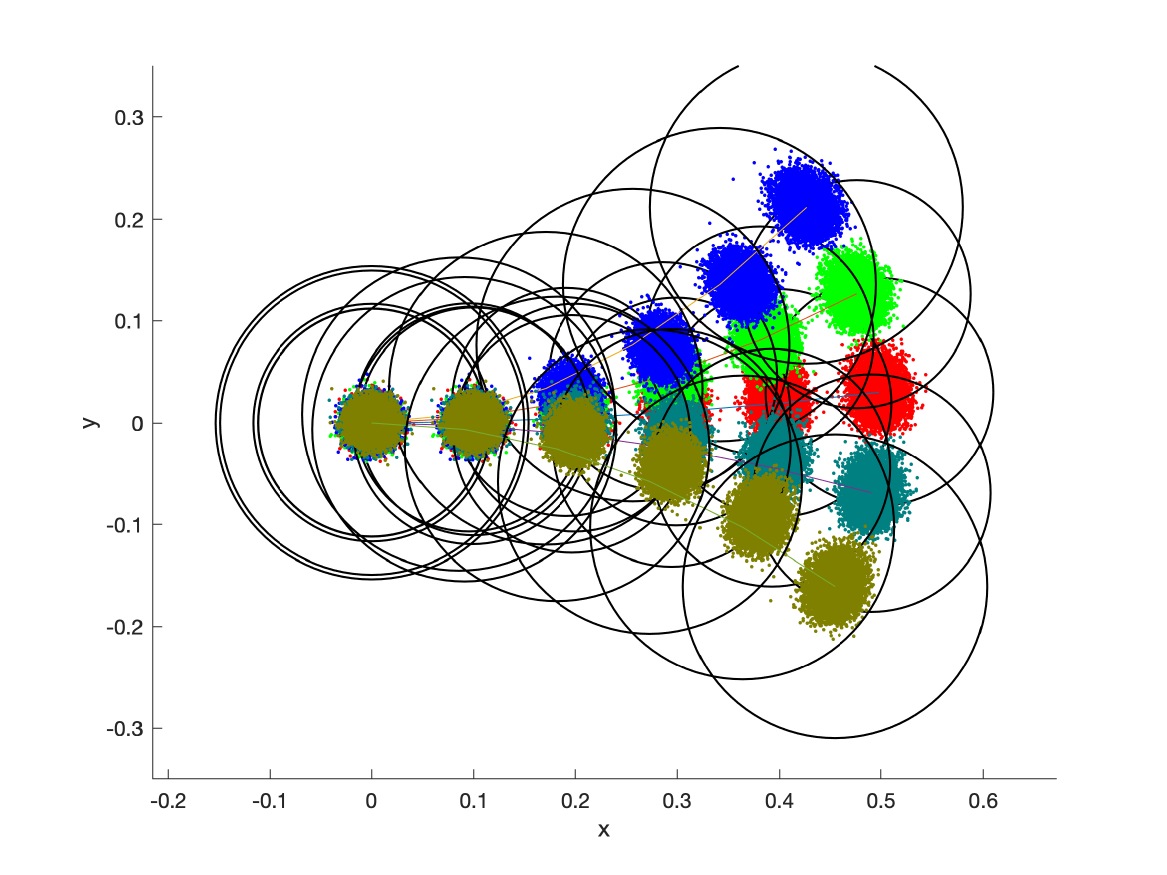}
    \caption{Motion primitives and their corresponding tubes. The top two plots are motion primitives and tubes for the underwater vehicle model. The bottom two plots are motion primitives and tubes for the ground vehicle model. In each plot, 10,000 samples are simulated forward using each motion primitive. Each black circle surrounding particles is the corresponding tube at that particular time step. In the left two plots, the tube sizes are computed using sampling, and in the right two plots, the tube sizes are computed using analytical methods. Although the tubes on the right are larger in size, they provide theoretical guarantees that at least 99.9\% of the system states stay inside the tube. }
    \label{fig:mps}
\end{figure}

\subsection{Tube Construction}\label{sec:method_4_tube_construction}
A tube along the polynomial nominal trajectory $\bar{\tbx}(t)$ in the form of 
\begin{align}\label{tube}
\mathcal{Q}_\phi(\bar{\tbx}(t)) =\{ \mathbf{x} \in \mathbb{R}^{n_x}:  (\mathbf{x}-\bar{\tbx}(t))^T{Q}_\phi(\mathbf{x}-\bar{\tbx}(t))\leq 1 \}  
\end{align}
for $t\in [0,1]$ is constructed so that the probability of system states staying inside the tube is $\geq (1-\Delta_{tube})$, where $\Delta_{tube} > 0$ is a small positive number, i.e.,
\begin{align}\label{eq:tube_constr}
    \text{Prob}(\tbx_t \in \mathcal{Q}_\phi(\bar{\tbx}(t))) \geq 1-\Delta_{tube}
\end{align}
Here the parameter to be determined is $\phi \in \Phi$, which controls the size of the tube. 
For example, in 2D the tube at time $t$ can be a disc 
\begin{align}\label{eq:quad_tube}
    \mathcal{Q}_\phi(\bar{\tbx}(t)) =\{ \mathbf{x} \in \mathbb{R}^{2}:  |(\mathbf{x}-\bar{\tbx}(t))|^2\leq r^2 \}
\end{align}
and the parameter is $\phi = r^{-1}$ for ${Q}_\phi = \phi * I_2$, where $I_2$ is the $2\times 2$ identity matrix.

Note that tube is continuous-time, while motion primitives are discrete-time. The discrete time steps in motion primitives correspond to $t = 0, {1 \over T}, \ldots, {T-1 \over T}, 1$, in the interval $t\in [0,1]$. 
We want to ensure that Inequality (\ref{eq:tube_constr}) holds for any $t = 0, {1 \over T}, \ldots, {T-1 \over T}, 1$. 
Clearly a larger tube size would allow more system states to stay inside the tube, but it would be more conservative for high level state space planning when we check the safety of the tube against obstacles. 
So we want the tube size to be tight.
For certain parameterizations of tubes, such as discs as in \Cref{eq:quad_tube}, we can use binary search to determine the min-sized tube, so that a little smaller in size would break Inequality (\ref{eq:tube_constr}).
In order for binary search to work, we need a procedure to verify if the tube of size $\phi$ satisfies Inequality (\ref{eq:tube_constr}) or not.
There are two ways -- one is sampling, and the other is an analytical method, which provides theoretical guarantees. 

\subsubsection{Method 1} Sampling. We sample $N$ states, and simulate forward using stochastic dynamics. Given a $\phi$, at each time step $k$ over the horizon $0, 1,\ldots, T$, the number of samples inside the tube $\mathcal{Q}_\phi(\bar{\tbx}(k/T))$ is $N_{in}(k)$. If $N_{in}(k)/N < 1 - \Delta_{tube}$ for some $k$, then the parameter $\phi$ is rejected. Otherwise $\phi$ is feasible. 
Different from the analytical method, the sampling method does not provide any theoretical guarantees, but it usually generates smaller sized tubes than the analytical method \Cref{fig:mps}.

\subsubsection{Method 2} Analytical method. The analytical method applies to systems whose future state moments can be calculated analytically.
This class of systems includes most robotic systems, such as vehicles and drones, as well as the examples in \cite{han2022non, jasour2021moment} and examples in this paper, with state-independent control inputs or certain state-dependent controls, which are to be shown in the following examples.
Given a $\phi$, at each time step $k$ over the horizon $0, 1,\ldots, T$, we want 
\begin{align}
    \text{Prob}(\tbx_k \not\in \mathcal{Q}_\phi(\bar{\tbx}(k/T))) \leq \Delta_{tube}
\end{align}
i.e.,
\begin{align} \label{eq:tube_size_analytical_2}
    \text{Prob}((\tbx_k-\bar{\tbx}(t))^T{Q}_\phi(\tbx_k-\bar{\tbx}(t))\geq 1, t=k/T) \leq \Delta_{tube}
\end{align}
The probability on the left-hand side of (\ref{eq:tube_size_analytical_2}) can be relaxed using concentration inequalities, such as Cantelli's inequality (\ref{eq:concentration_ineq}), and the probability inequality (\ref{eq:tube_size_analytical_2}) can be converted to an inequality involving moments of $\tbx_k$, as in \Cref{sec:method_1}.
If the moment inequalities are satisfied for all $k$, then $\phi$ is feasible, otherwise $\phi$ is rejected.
In contrast to sampling, the analytical method find the min-sized tube that guarantees Inequality (\ref{eq:tube_constr}).

\textbf{Example 1.} Consider the underwater robot model given by
\begin{equation}\label{eq:dyn1}
\begin{aligned}
    &x_{t+1} = x_{t} + \Delta T (v_t + {\omega_v}_t) \cos(\theta_t + {\omega_{\theta}}_t)\\
    &y_{t+1} = y_t + \Delta T (v_t + {\omega_v}_t) \sin(\theta_t + {\omega_{\theta}}_t)
\end{aligned}
\end{equation}
where $(x_t,y_t)$ is the 2D position at time $t$, and the discrete time interval $\Delta T = 0.1$. The noise terms ${\omega_v}_t$ and ${\omega_{\theta}}_t$ have uniform distribution on $[-0.1, 0.1]$ at any time $t$.
We generate 5 motion primitives with horizon $T = 5$, and fit each of them with a quadratic function of the form 
\begin{equation}\label{tube_traj_quad}
\begin{aligned}
    & \bar{\tbx}(t) = (x(t), y(t))\\
    & x(t) = \theta_1 t\\
    & y(t) = \theta_2 x(t)^2
\end{aligned}
\end{equation}
for $t\in [0,1]$, where $\theta = (\theta_1, \theta_2)$ is the parameter.
We choose tube to be discs as in \Cref{eq:quad_tube}, and we use binary search to find the min-sized tube. 
The control inputs of the motion primitives are state independent and hence the analytical method applies.
We use both sampling (\Cref{fig:mps} top left) and the analytical method (\Cref{fig:mps} top right) to construct the tube for each motion primitive.
The analytical method guarantees that at least $99.9\%$ of the system states stay inside the tube.

\textbf{Example 2.} Consider the ground vehicle model whose dynamics is in the form of 
\begin{equation}\label{eq:dyn2}
\begin{aligned}
    &x_{t+1} = x_{t} + \Delta T v_t  \cos(\theta_t)\\
    &y_{t+1} = y_t + \Delta T v_t \sin(\theta_t)\\
    &v_{t+1} = v_t + \Delta T (a_t + {\omega_{v_t}})\\
    &\theta_{t+1} = \theta_t + \Delta T (u_t + \omega_{\theta_t})
\end{aligned}
\end{equation}
where the state is $(x_t, y_t, v_t, \theta_t)$, $(x_t, y_t)$ is the 2D position at time $t$, $v_t$ is the velocity at time $t$, and $\theta_t$ is the angle at time $t$. The discrete time interval is $\Delta T = 0.1$. The noise $\omega_{v_t}$ has normal distribution with mean 0 and variance $0.09$, while $\omega_{\theta_t} \sim 3B$, where $B$ has Beta distribution with parameters $(1, 3)$ over $[0, 1]$. 
The initial distribution of $x_0$ and $y_0$ are both normal distribution with mean 0 and variance $0.01^2$, while the initial deterministic states of $v_0$ and $\theta_0$ are $1$ and 0, respectively.
Similar to Example 1, we generate 5 motion primitives with horizon $T = 5$, and fit each primitive with a quadratic nominal trajectory in the form of \Cref{tube_traj_quad}. 
The tubes are discs as in \Cref{eq:quad_tube}.
The control inputs of the motion primitives are state dependent, and they are to track certain nominal velocities $\bar{v}_t$ and nominal angles $\bar{\theta}_t$, i.e., the control inputs are $a_t = (\bar{v}_{t+1} - v_t)/\Delta T$, $u_t = (\bar{\theta}_{t+1} - \theta_t)/\Delta T$.
The dynamics for $v$ and $\theta$ essentially becomes $v_{t+1} = \bar{v}_{t+1} + \Delta T{\omega_{v_t}}$, and $\theta_{t+1} = \bar{\theta}_{t+1} + \Delta T \omega_{\theta_t}$.
For this system, the moments of system states can be calculated analytically.
We use both sampling (\Cref{fig:mps} bottom left) and the analytical method (\Cref{fig:mps} bottom right) to construct the tube for each motion primitive.
The analytical method guarantees that at least $99.9\%$ of the system states stay inside the tube.

\subsection{Online Execution}\label{sec:method_5_online_execution}
So far we have constructed various motion primitives and their corresponding tubes offline. 
During online execution, the system, based on the current state, selects a motion primitive that is risk bounded, verified via SOS programming in \Cref{sec:method_2_tube_verification}, with respect to its surrounding obstacles. 
The system executes the first one or few steps of the motion primitive and re-plans based on the new system state, much like model predictive control (MPC) style planning. 
One planning cycle consists of selecting a risk-bounded motion primitive based on the current state and executing the first one or few steps of the motion primitive.

The most naive implementation for selecting a risk-bounded motion primitive is to verify all tubes against all obstacles, and select one of the tubes that are risk bounded with respect to all obstacles, and execute its corresponding motion primitive.
Since online computation resources are limited, we can optimize the implementation in a number of ways:
\begin{itemize}
    \item If the motion primitives are ordered from left to right, viewed in the direction of the current velocity, then binary search algorithm can be used to look for feasible tubes against a particular obstacle.
    \item We only check the safety of the tubes against obstacles within certain distance of the system. Obstacles far away can clearly be ignored.
    \item Use lower order polynomials, such as quadratics, circles, and ellipses, to represent the obstacles would make verification faster, though they might be more conservative. 
\end{itemize}
Finally, instead of picking feasible motion primitives randomly, we can use an objective to rank all feasible motion primitives and the one with the highest score will be selected.

\subsection{Theoretical Guarantees and Summary}
\noindent\textbf{Theoretical Guarantees of Bounded Risk.}
During offline planning, we generate motion primitives, fit polynomial nominal trajectories, and build tight tubes so that if the tube starts from the current state, then the probability of system states staying inside the tube $\mathcal{Q}(\bar{\tbx}(t))$ with $t\in [0,1]$, for any $t$ in the planning horizon of the motion primitive, is at least $1-\Delta_{tube}$, i.e., for $t = 0, {1\over T},\ldots,1$,
\begin{align}\label{eq:theoretical_guarantee_eq1}
\text{Prob}(\tbx_t \in \mathcal{Q}(\bar{\tbx}(t))) \geq 1-\Delta_{tube}.
\end{align}
We convert the uncertain environment into risk contours so that we can check if a tube $\mathcal{Q}(\bar{\tbx}(t))$ is in the risk-bounded set $\hat{\mathcal{C}}_{r}^{\Delta}(t))$ using SOS programming. If a tube $\mathcal{Q}(\bar{\tbx}(t))$ is inside the risk-bounded set $\hat{\mathcal{C}}_{r}^{\Delta}(t))$, then the probability of the system states inside the tube colliding with any obstacle $\mathcal{O}$ is bounded by $\leq \Delta_{o}$, i.e., 
\begin{align}\label{eq:theoretical_guarantee_eq2}
\text{Prob}(\tbx_t \in \mathcal{O} |\tbx_t \in \mathcal{Q}(\bar{\tbx}(t)) \subseteq \hat{\mathcal{C}}_{r}^{\Delta}(t)) \leq \Delta_{o}.    
\end{align}
During online execution, we select a motion primitive whose tube is inside the risk-bounded set $\hat{\mathcal{C}}_{r}^{\Delta}(t)$. Therefore, the probability of system states colliding with any obstacle is bounded by $\Delta_o +\Delta_{tube}$, since
\begin{align}\label{eq:bound_one_cycle}
    & \quad \text{Prob}(\tbx_t \in \mathcal{O} | \mathcal{Q}(\bar{\tbx}(t)) \subseteq \hat{\mathcal{C}}_{r}^{\Delta}(t)) \nonumber\\
    & \leq  \text{Prob}(\tbx_t \in \mathcal{O} |\tbx_t \in \mathcal{Q}(\bar{\tbx}(t)) \subseteq \hat{\mathcal{C}}_{r}^{\Delta}(t)) \text{Prob}(\tbx_t \in \mathcal{Q}(\bar{\tbx}(t))) \nonumber\\
    & \quad + \text{Prob}(\tbx_t \not\in \mathcal{Q}(\bar{\tbx}(t))) \quad  (\text{from Law of Total Probability}) \nonumber\\ 
    &\leq {\Delta_o +\Delta_{tube} } \quad\quad\quad\quad\ \  (\text{from } (\ref{eq:theoretical_guarantee_eq1}), (\ref{eq:theoretical_guarantee_eq2}))
\end{align}

The bound in (\ref{eq:bound_one_cycle}) is only for one tube, or one planning cycle. Suppose the system reaches the goal in $N$ planning cycles. Then there are $N$ tubes connected consecutively, paving the path from the initial position to the goal region.  
Let $A_N$ denote the event that the system stays in the tubes over all the $N$ planning cycles. 
Let $B_N$ denote the event that the system collides with any obstacle over all the $N$ planning cycles.
Then
\begin{align}
    \text{Prob}(A_N) = \text{Prob}(\tbx_t \in \mathcal{Q}(\bar{\tbx}(t)))^N \geq (1-\Delta_{tube})^N.
\end{align}
By the same reasoning as in (\ref{eq:bound_one_cycle}),
\begin{align}
    \text{Prob}(B_N) &\leq \Delta_o + (1 - (1-\Delta_{tube})^N) \label{eq:N_cycle_bound_1}\\
    &\leq \Delta_o + N \Delta_{tube} \label{eq:N_cycle_bound_2}
\end{align}
Both (\ref{eq:N_cycle_bound_1}) and (\ref{eq:N_cycle_bound_2}) are bounds on the probability of colliding with any obstacle over the entire $N$ planning cycle.
When $\Delta_{tube}$ is very small, e.g., $0.001$ in our examples, both bounds are close to each other. 

\noindent \textbf{User-side Risk Allocation.}
Suppose $\Delta$ is the total risk  the user considers.
The user can first estimate an upper bound $M$ on the number of total planning cycles, and then choose $\Delta_o$ and $\Delta_{tube}$ so that 
\begin{align}
    \Delta_o + M \Delta_{tube} \leq \Delta. \label{eq:risk_allocation}
\end{align}
Suppose there are $N$ actual planning cycles, where $N\leq M$.
Then the risk is bounded by $\Delta$, i.e.,
\begin{align}
    \text{Prob}(B_N) \leq \Delta_o + N \Delta_{tube}
    \leq \Delta_o + M \Delta_{tube} \leq \Delta.
\end{align}

For example, suppose we are given a total risk $\Delta = 0.1$, and we estimate the number of total planning cycles is bounded by $M = 100$. 
Then we can split the total risk into halves, choosing $\Delta_o = M \Delta_{tube} = {1\over 2} \Delta = 0.05$, and hence $\Delta_{tube} = 0.0005$. 
In our experiments, varying $\Delta_o$ and $\Delta_{tube}$ does not affect conservativeness or efficiency too much. 

\noindent \textbf{Algorithm Summary.}
We summarize the algorithm in \Cref{algo}.
Note that Line 5 says that we generate risk contours of the uncertain environment online at each time step. 
This can vary depending on the robot platform and the environment. 
If the environment is static, then risk contours can be generated offline.
If the environment is highly dynamic, such as a traffic scene, and if the robot is an autonomous driving car with strong computing power, then the prediction module of the car predicts the future trajectories of surrounding vehicles at high frequency \cite{gu2021densetnt, lu2022kemp} and risk contours can be form at high frequency, too.

\begin{algorithm}
\nonl \textbf{\textit{Offline}}:\\
Given a total risk $\Delta$, estimate an upper bound $M$ of the total planning cycle, and choose $\Delta_o$ and $\Delta_{tube}$ satisfying \Cref{eq:risk_allocation}.\\
Design motion primitives (\Cref{sec:method_3_motion_primitives}).\\
Build tubes for the motion primitives (\Cref{sec:method_4_tube_construction}).\\ 
\nonl \textbf{\textit{Online}}:\\
For each time step until goal is reached:\\
(1) Generate risk contours of the uncertain environment (\Cref{sec:method_1}).\\
(2) Check tubes against risk contours (\Cref{sec:method_2_tube_verification}).\\
(3) Rank the tubes that are safe using a user-defined objective, and choose the one with the highest score (\Cref{sec:method_5_online_execution}).\\
(4) Execute the first one or few time steps of the chosen motion primitive (\Cref{sec:method_5_online_execution}). 

\caption{Real-Time Tube Based Non-Gaussian Risk Bounded Motion Planning}
\label{algo}
\end{algorithm}

\section{Experiments}
\begin{figure}[t!]
    \centering
    \includegraphics[width=0.23\textwidth]{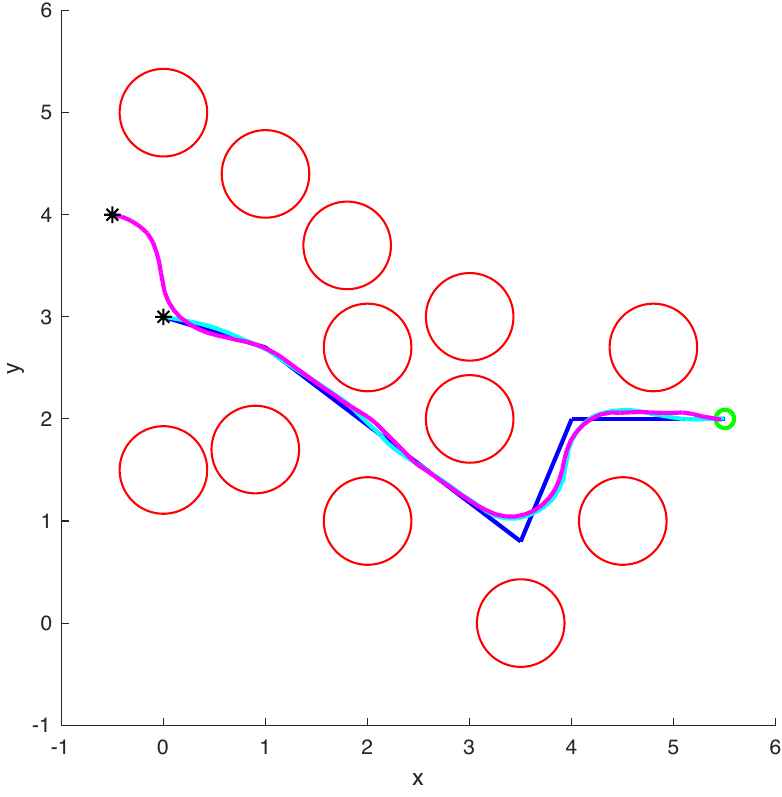}
    \includegraphics[width=0.23\textwidth]{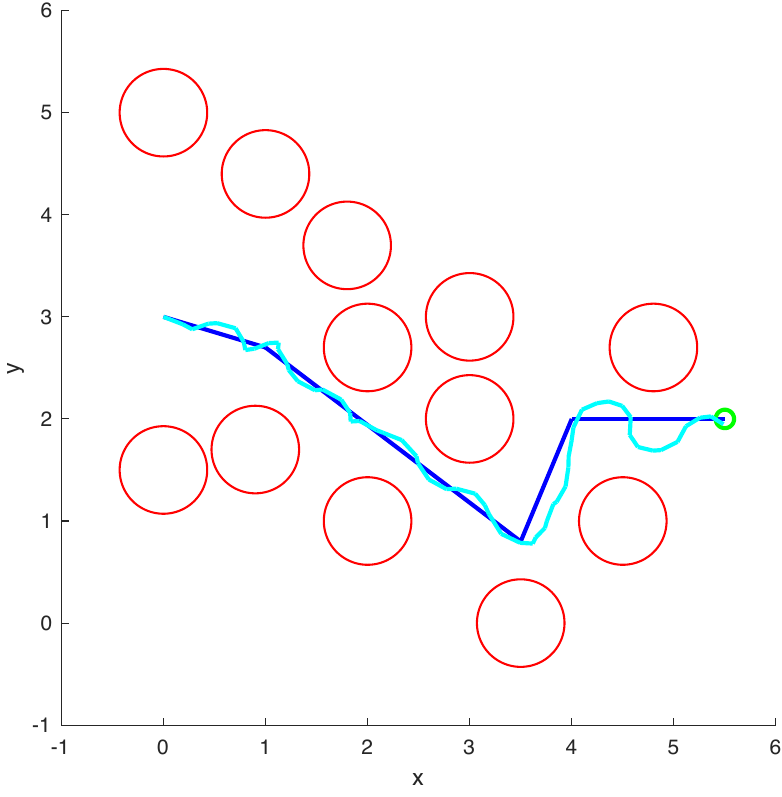}
    \caption{Underwater Robot in Clustered Environment. Left: Our method. Right: CC-RRT. The red curves are boundaries of obstacles' risk contours, which are 4th order polynomials. The blue piecewise-linear trajectory is the general path given by the high-level planner. The green circle represents the goal region. The plot on the left shows two trajectories given by our method, colored in magenta and cyan, starting from two different initial positions, marked by two black stars, and reaching the goal in the end. The plot on the right shows a cyan trajectory given by CC-RRT.}
    \label{fig:exp1}
\end{figure}

\begin{figure}
    \centering
    \includegraphics[width=0.35\textwidth]{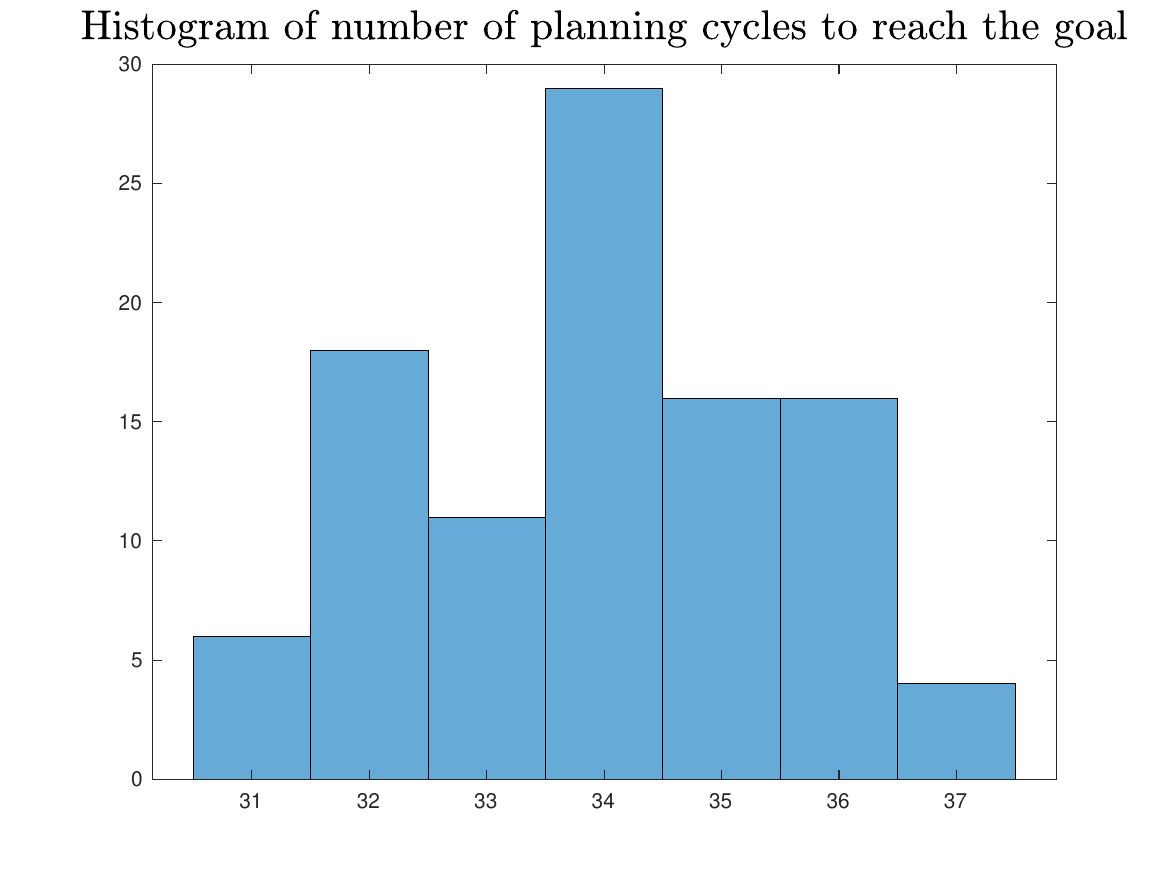}
    \caption{Histogram of the number of planning cycles to reach the goal from 100 initial states.}
    \label{fig:exp1_hist}
\end{figure}
\subsection{Underwater Robot in Clustered Environment}
In this example, we consider the underwater vehicle whose dynamics is given by \Cref{eq:dyn1}. In the environment there are multiple static obstacles, and we approximate them, in the spirit of \Cref{sec:method_5_online_execution}, in the form of
\begin{align*}
    \{(x,y) \in \R^2: (x - x_i)^2 + (y - y_i)^2 \leq w^2\}
\end{align*}
where $(x_i, y_i)$ is the center for the $i$-th obstacle, and $w$ is a random variable with uniform distribution on $[0.3, 0.4]$.
The risk contours of the obstacles are 4th order polynomials in 2 variables and we plotted only their boundaries in (\Cref{fig:exp1}).
We want the total risk level to be $\Delta = 0.2$.
We estimate the total number of planning cycles is bounded by 100. 
So we choose risk contours with the risk level of $\Delta_o = 0.1$, and choose $\Delta_{tube} = 0.001$. 
We design motion primitives using the analytical method.
The goal region is given by $\{(x,y) \in \R^2: (x-x_g)^2 + (y-y_g)^2 \leq r_g^2\}$, where $(x_g, y_g) = (5.5,2)$ and $r_g = 0.09$.
The high-level planner provides a general path represented by the blue piecewise linear trajectory in \Cref{fig:exp1}.
The objective to rank the feasible motion primitives is the sum of squared distance between expected future states of the motion primitive and the path given by the high-level planner. The feasible motion primitive that minimizes the objective is selected.

We assume the initial state has uniform distribution on the square with side length 1 centered at the point $(0,3)$.
We sample 100 initial states and apply our method to those samples.
The system re-plans every 2 time steps.
All samples reach the goal while staying inside the tube. 
The success rate is 100\%.
We plotted the histogram of the number of planning cycles to reach the goal in \Cref{fig:exp1_hist}.
The max number of planning cycles is 37, which is less than our estimate of 100.
This verifies that our estimate is indeed a good upper bound. 

In \Cref{fig:exp1} Left, we plotted the trajectory starting from two initial positions, $(0,3)$ and $(-0.5,4)$, both with initial velocity $(1,0)$.  
The initial state $(-0.5,4)$ is actually outside of the initial distribution. 
The trajectory starting from $(0,3)$ finishes in $N_1 = 34$ planning cycles, and hence the risk is bounded by $\Delta_o +N_1\Delta_{tube} =0.134$, or $\Delta_o +(1-(1-\Delta_{tube})^{N_1}) =0.1335$.
The trajectory starting from $(-0.5,4)$ finishes in $N_2 = 40$ planning cycles, and hence the risk is bounded by $\Delta_o +N_2\Delta_{tube} =0.14$, or $\Delta_o +(1-(1-\Delta_{tube})^{N_2}) =0.1393$.

\begin{figure}
    \centering
    \includegraphics[width=0.25\textwidth]{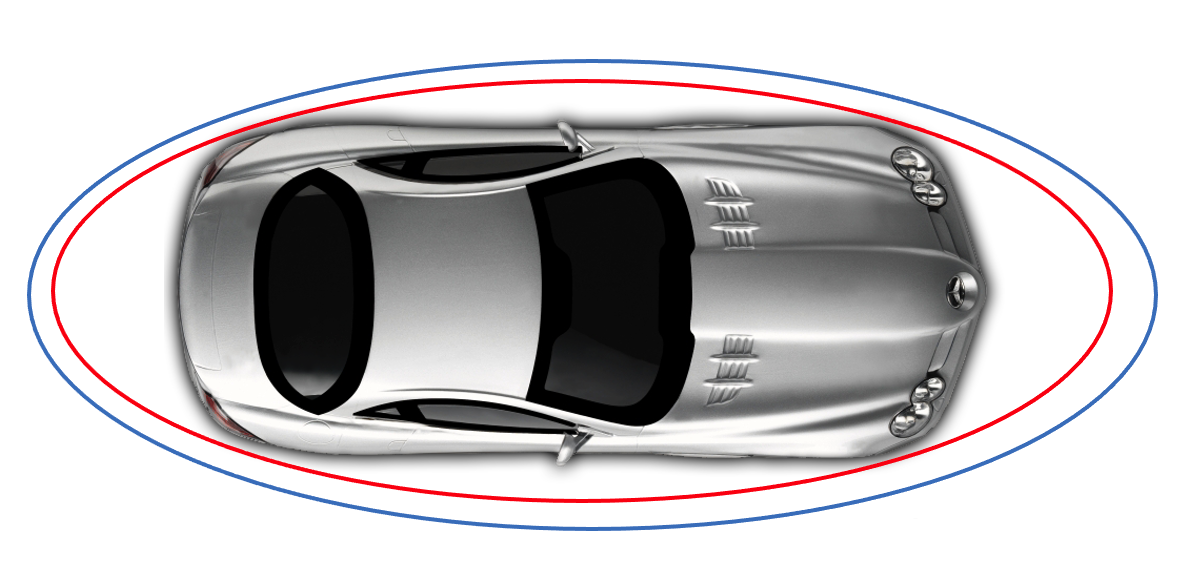}
    \caption{Outer approximation of the risk contour of a vehicle. The red curve is the boundary of the risk contour, which is a 4th order polynomial in 3 variables. The blue ellipse is an outer approximation of the risk contour, which is a quadratic function in 3 variables.}
    \label{fig:outer_approx_car}
\end{figure}

\begin{figure}[t!]
    \centering
    \includegraphics[width=0.15\textwidth]{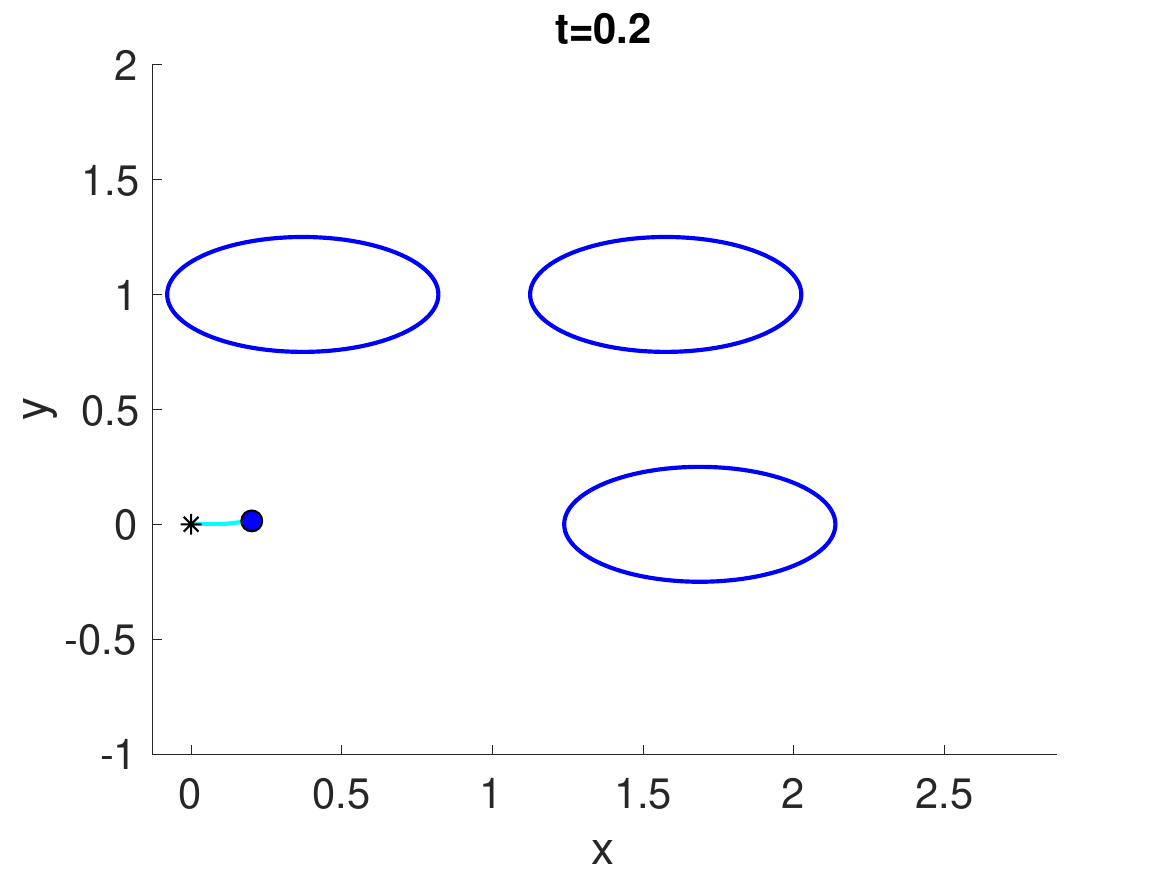}\ 
    \includegraphics[width=0.15\textwidth]{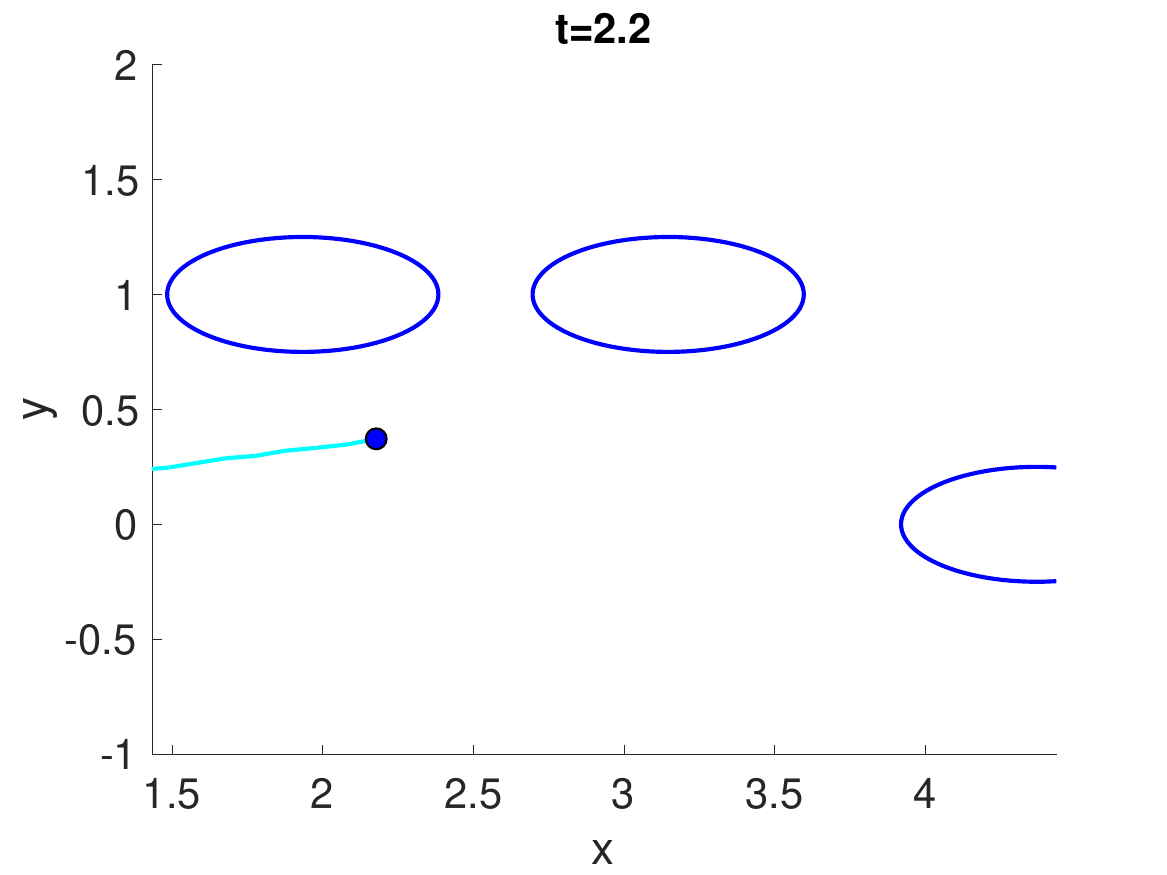}\
    \includegraphics[width=0.15\textwidth]{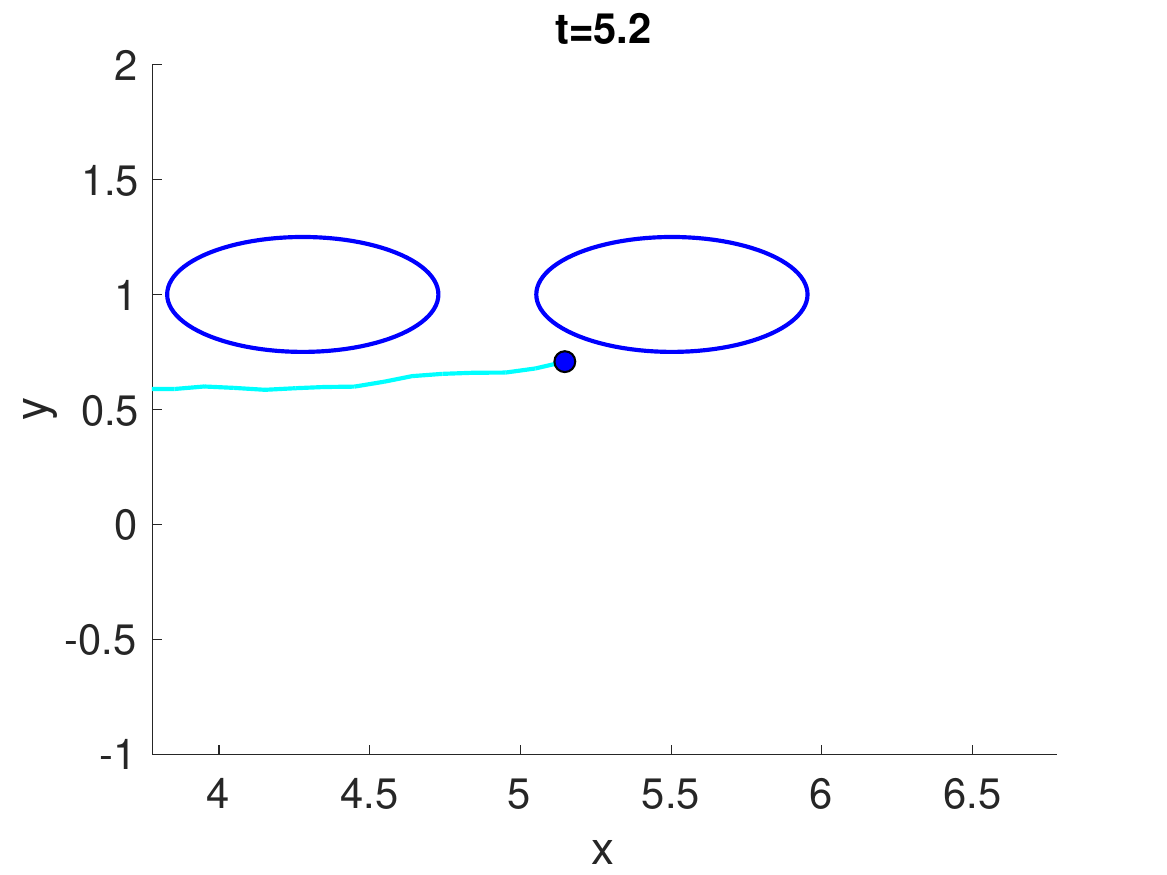}
    
    \includegraphics[width=0.15\textwidth]{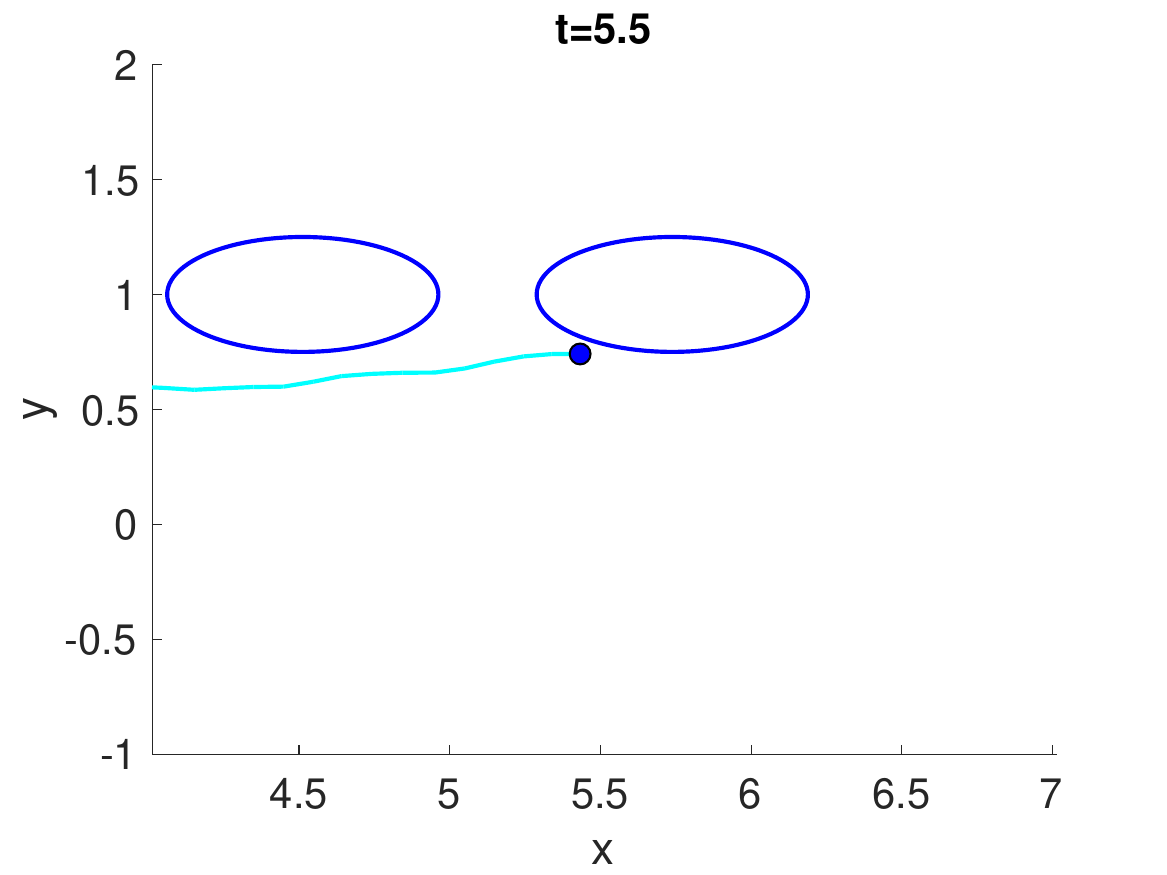}\
    \includegraphics[width=0.15\textwidth]{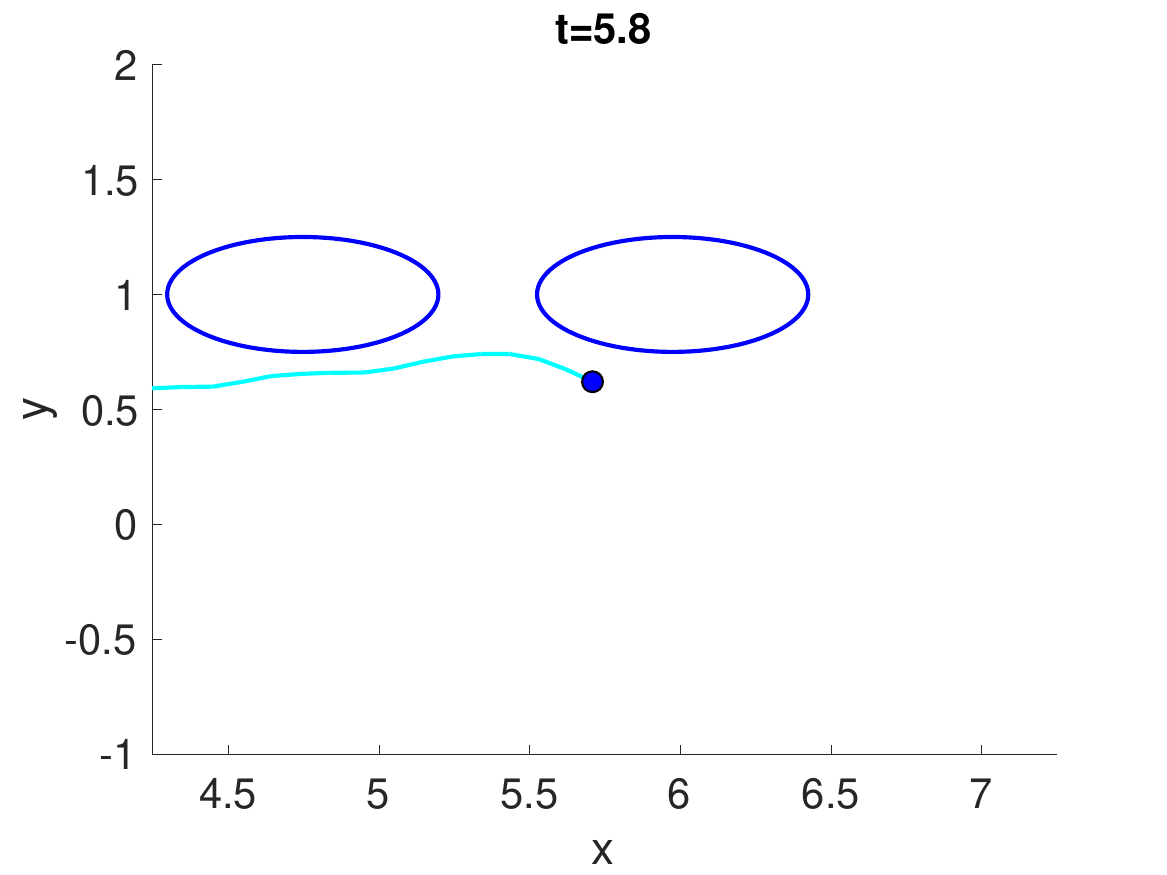}\
    \includegraphics[width=0.15\textwidth]{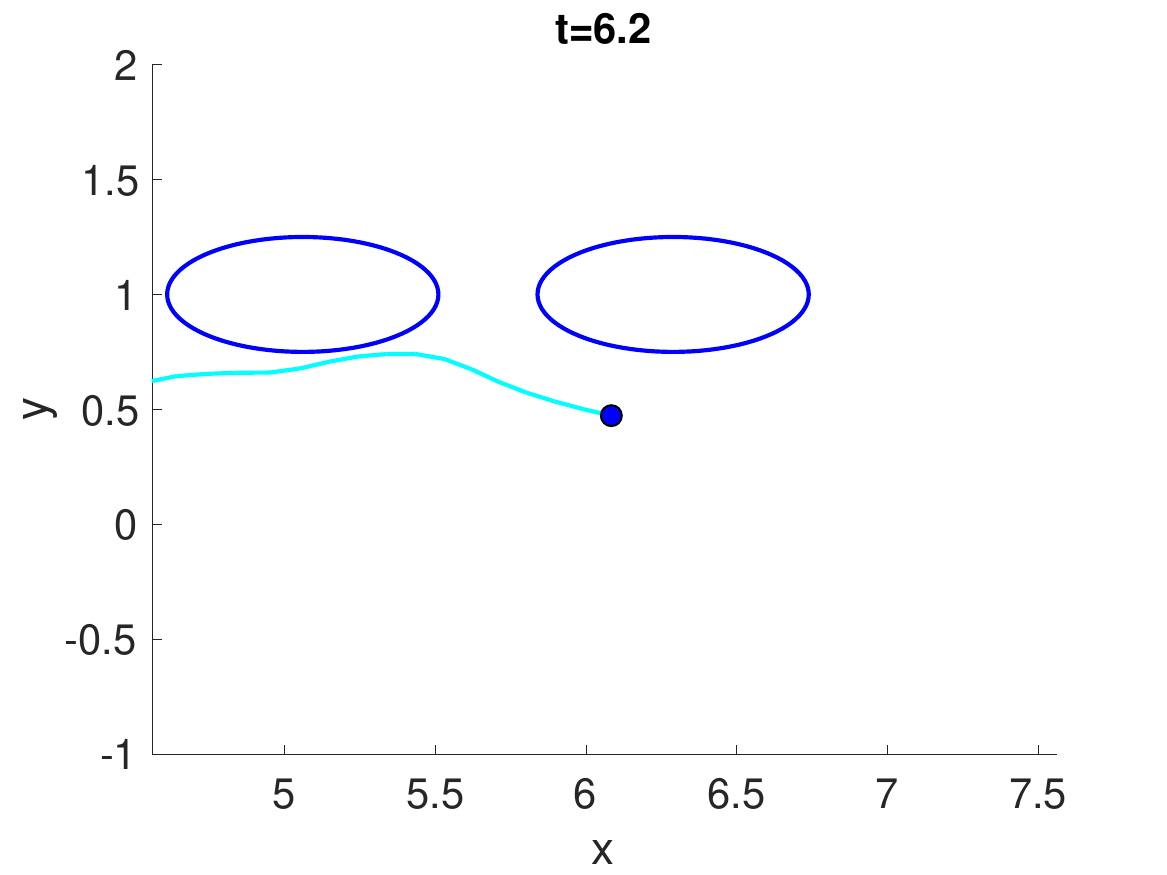}
    
    \ \includegraphics[width=0.15\textwidth]{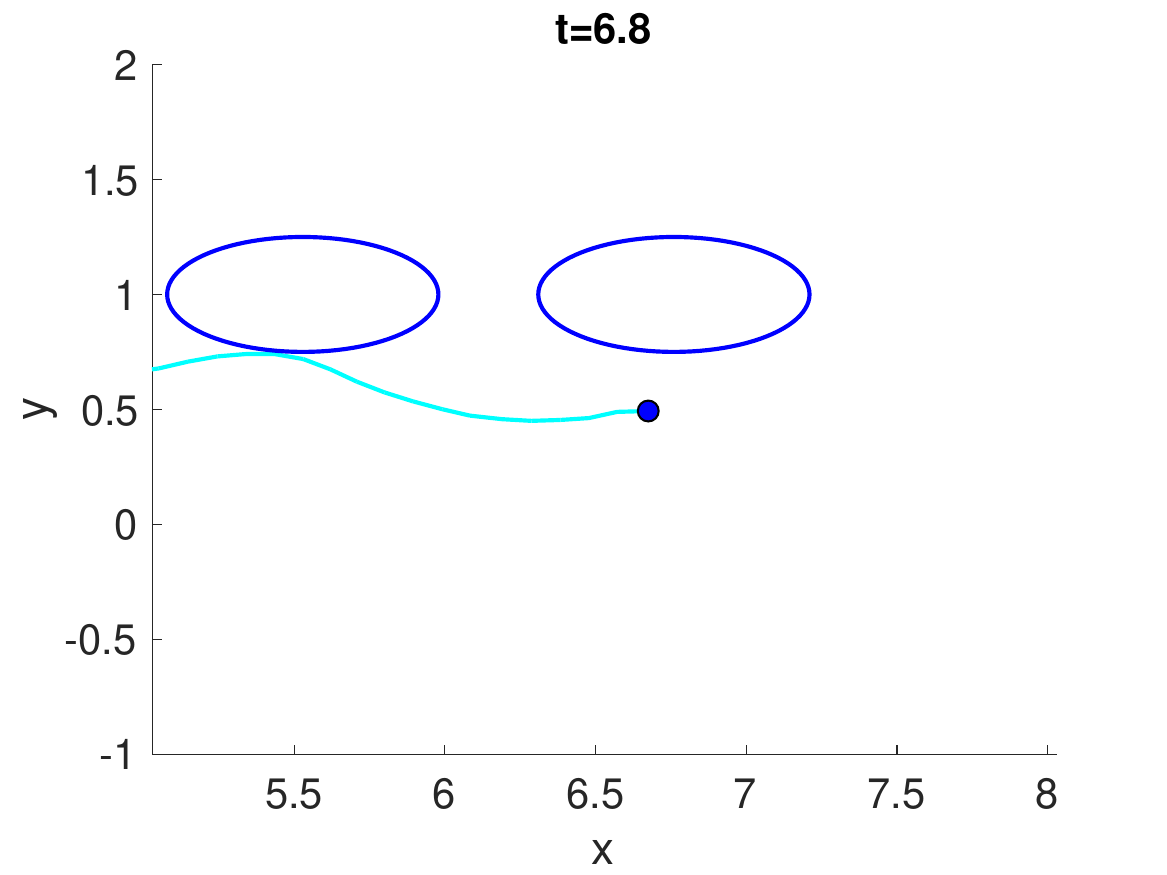}\ 
    \includegraphics[width=0.15\textwidth]{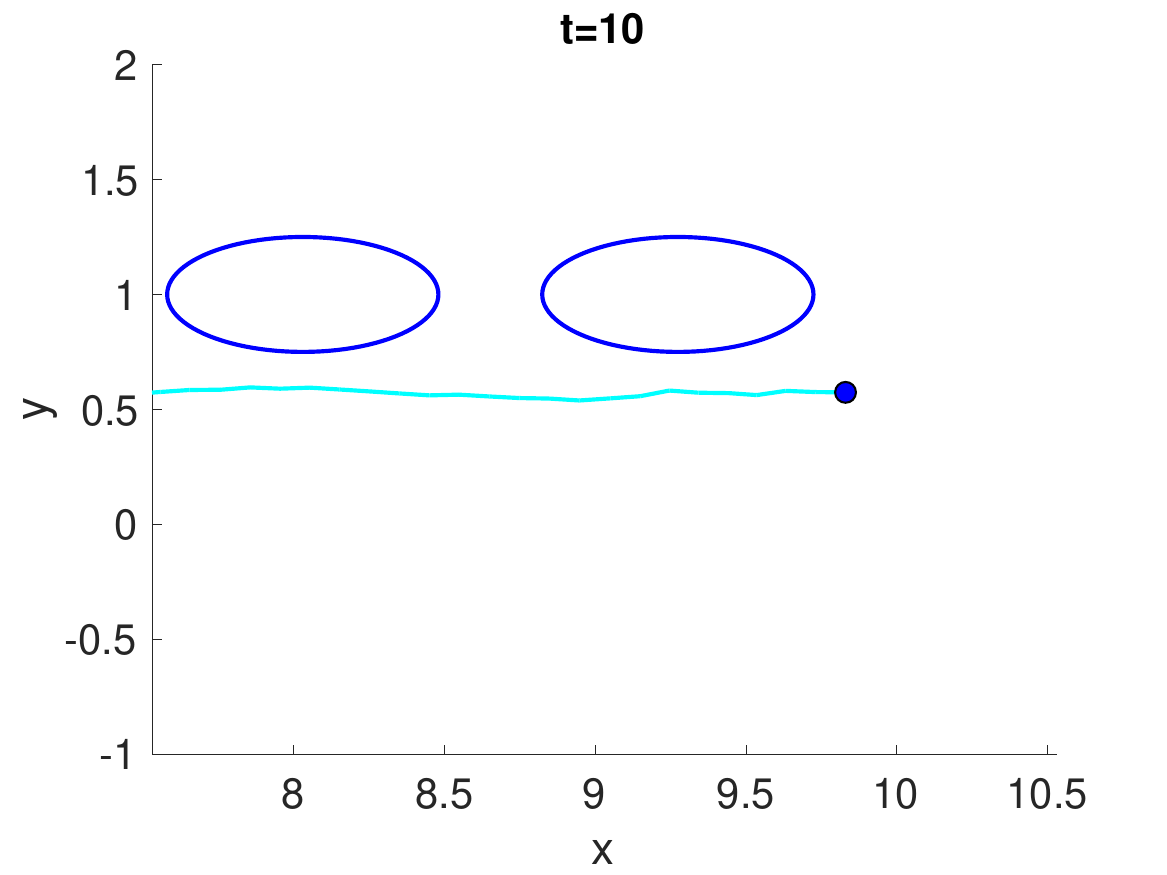}\
    \includegraphics[width=0.15\textwidth]{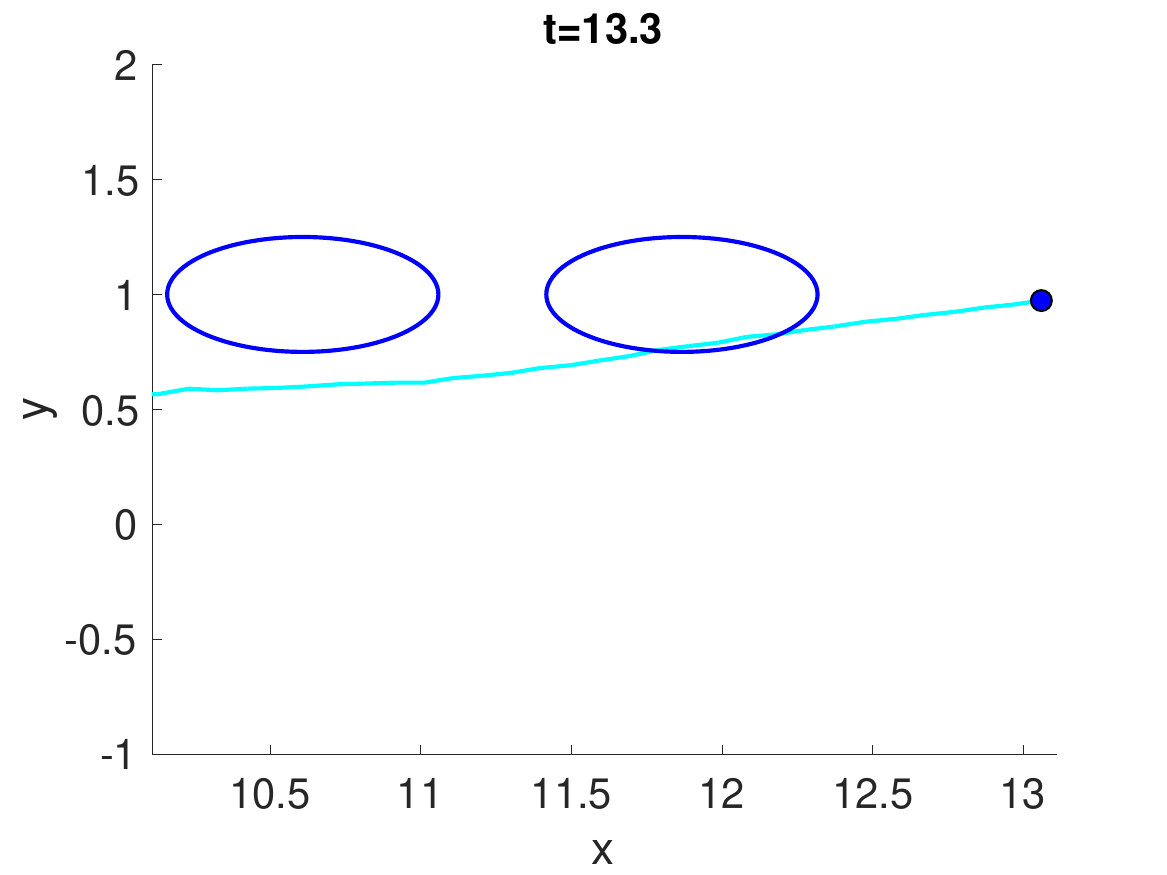}
    \caption{Ground vehicle changing lanes. The plots are the snapshots of the state space at time $t = 0.2, 2.2, 5.2, 5.5, 5.8, 6.2, 6.8, 10$, and $13.3$. 
    All vehicles are moving to the right.
    Blue ellipses are risk contours of the surrounding vehicles. The camera frame is relatively static to the top left vehicle, and hence the top left vehicle looks static across the images. 
    In this scene, two vehicles on the top have almost the same velocity.
    The cyan curve represents the trajectory the system has taken so far. The blue dot at the right end of the cyan curve is the current system position. 
    The system reaches the goal at time 13.3.}
    \label{fig:exp2}
\end{figure}

\noindent\textbf{Comparison with CC-RRT.} 
We compare our method with Chance-Constrained RRT (CC-RRT) method proposed in \cite{luders2010chance}. 
In their setting, the system dynamics is linear and has Gaussian noise. 
The initial position of the system has Gaussian distribution.
The obstacles are convex polyhedra.

We apply CC-RRT to the clustered environment and a trajectory is plotted as a cyan curve in \Cref{fig:exp1} Right. The system dynamics is simplified as 
\begin{align*}
x_{t+1} &= x_t + \Delta T * v_{x,t} + w_{x,t}\\
y_{t+1} &= y_t + \Delta T * v_{y,t} + w_{y,t}
\end{align*}
where $w_{x,t}$ and $w_{y,t}$ are Gaussian distributions with mean 0 and standard deviation $0.02$.
Similar to our online planning method, we do MPC style planning and at each planning cycle, we do RRT search and the safe candidates are ranked by an objective function. 
The objective function is the distance of the future state to the path given by the high level planner, plus a weighted distance of the future state to the goal.
The safe candidate with the lowest score is selected and executed.

Our method has several advantages over CC-RRT: (i) Our method can bound the risk using tubes over the continuous state space, while CC-RRT can only check safety at certain discrete points via sampling. (ii) Our method can work with nonlinear systems, while CC-RRT works with linear systems. (iii) Our method can deal with more general probabilistic distributions, while CC-RRT only works with Gaussian distribution. (iv) Our method can deal with more general obstacles, and does not assume the obstacles are convex, though using simple convex obstacles would speed up our method. 

\begin{figure}
    \centering
    \includegraphics[width=0.35\textwidth]{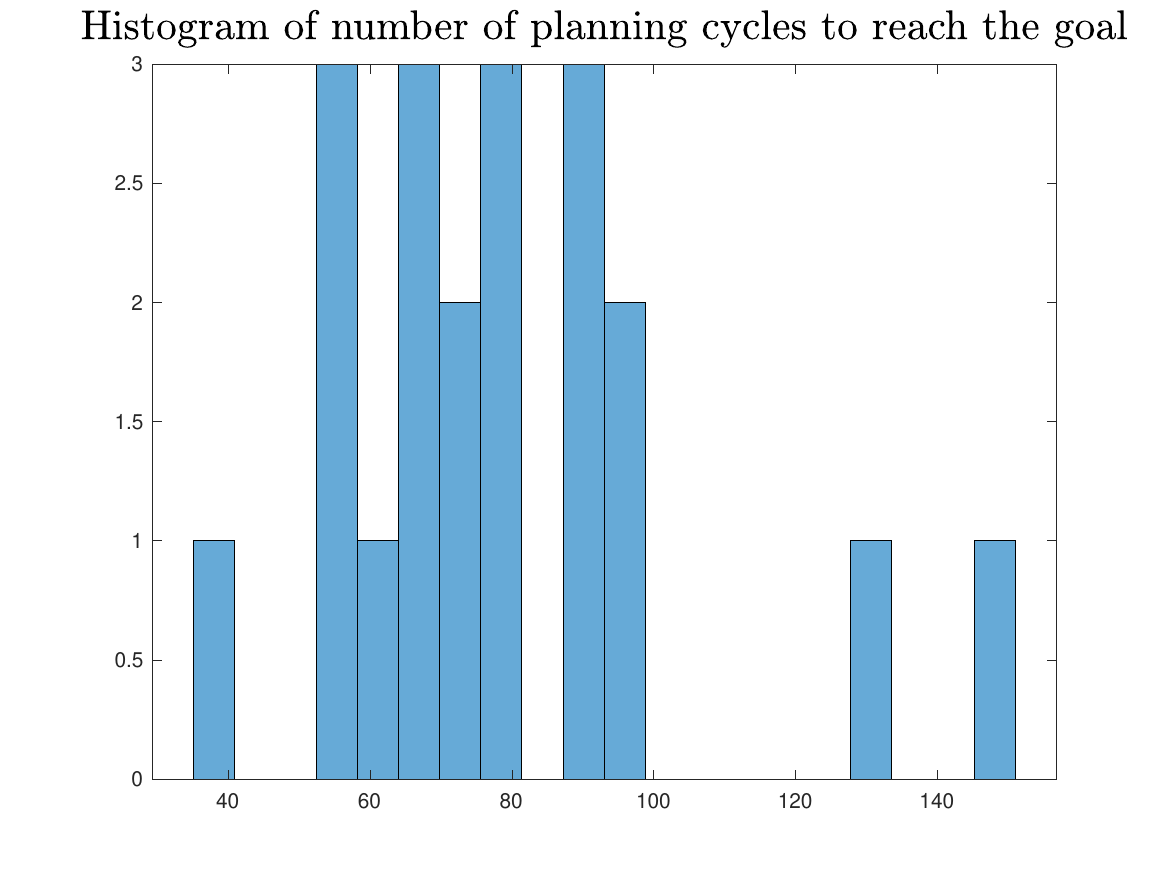}
    \caption{Histogram of the number of planning cycles to reach the goal across 20 scenes.}
    \label{fig:exp2_hist}
\end{figure}

\subsection{Vehicle Changing Lanes}
In this example, we consider the ground vehicle whose dynamics is given by \Cref{eq:dyn2}. 
We randomly generate 20 scenes.
In each scene, there are three uncertain vehicles moving towards the right. 
Two are moving on the top lane, and one is moving on the bottom lane.
The initial positions and velocities of the vehicles vary across the 20 scenes. 
Each vehicle has the form 
\begin{align*}
    \{(x,y) \in \R^2: (x - x_i)^2 + 4(y - y_i)^2 \leq w^2\}
\end{align*}
where $w$ has uniform distribution over $[0.3, 0.4]$. 
The system starts from the position $(0,0)$ moving with initial velocity $(1,0)$, and the goal is to switch lanes from $y = 0$ to $y = 1$.

We are given a total risk level of 0.3.
We estimate an upper bound on the number of planning cycles is 200.
We use risk contours with the risk level of $\Delta_o = 0.1$ and $\Delta_{tube} = 0.001$. 
The risk contour is a polynomial of order 4. 
We outer approximate the risk contour by an ellipse, which is a quadratic function, reducing the order of the risk contour to 2 (\Cref{fig:outer_approx_car}).
We build motion primitives using the analytical method.
There is no high level planner in this example.
Instead there is an cost function $(y - 1)^2 + 10 \theta^2 + 10^7 * I(|\theta - \pi/6| > 0)$ to minimize, where $y$ and $\theta$ are the states at the last time step of the motion primitive, and $I$ is the indicator function.
So a motion primitive that approaches the lane $y=1$ while keeping $\theta$ close to 0 is preferable.  
The task is accomplished once $y$ is close to 1 and $\theta$ is close to 0. 
The system re-plans at every time step.

Among the 20 scenes, the max number of planning cycles is 150 (\Cref{fig:exp2_hist}), which is less than our estimate of 200, and hence the total risk of 0.3 is guaranteed. 
We plotted an example trajectory at several different time steps in \Cref{fig:exp2}.
In this example, $N = 133$, and hence the entire trajectory has bounded risk of $\Delta_o +N\Delta_{tube} =0.233$ or $\Delta_o +(1-(1-\Delta_{tube})^{N}) =0.2246$.

\section{Conclusion and Future Work}
We have presented a real-time tube-based motion planning approach for stochastic nonlinear systems in uncertain environments via motion primitives. 
Our approach works for long-term tasks, which trajectory optimization methods, such as the one in \cite{han2022non}, cannot be directly applied.
Our approach guarantees the probability of system states colliding with any obstacle is bounded above.
Our approach is very practical and can be deployed on various robotics systems. Future work includes implementation on real robots and autonomous driving cars.

\bibliographystyle{IEEEtran}
\bibliography{references}

\end{document}